\documentclass[preprint]{elsarticle}
\usepackage{lineno, hyperref}
\modulolinenumbers[5]
\hypersetup{pdfauthor=author}

\usepackage[utf8]{inputenc}
\usepackage{graphicx}
\usepackage{amsfonts}
\usepackage{amsmath}
\usepackage{algorithm}
\usepackage{xcolor}
\usepackage[noend]{algpseudocode}
\usepackage{makecell}

\newcommand{\mean}{\texttt{mean}}
\newcommand{\colmean}{\texttt{colmean}}
\newcommand{\var}{\texttt{var}}
\newcommand{\imp}{\texttt{imp}}
\newcolumntype{R}[1]{>{\raggedleft\let\newline\\\arraybackslash\hspace{0pt}}m{#1}}

\journal{Knowledge-Based Systems}
\begin{document}

\begin{frontmatter}

\title{Oblique Predictive Clustering Trees}

\author[IJS,MPS]{Toma\v{z} Stepi\v{s}nik\corref{cor1}}
\ead{tomaz.stepisnik@ijs.si}
\author[IJS,MPS]{Dragi Kocev}
\ead{dragi.kocev@ijs.si}

\address[IJS]{Department of Knowledge Technologies, Jo\v{z}ef Stefan Institute, Ljubljana, Slovenia}
\address[MPS]{Jo\v{z}ef Stefan International Postgraduate School, Ljubljana, Slovenia}
\cortext[cor1]{Corresponding author}

\begin{abstract}
Predictive clustering trees (PCTs) are a well-established generalization of standard decision trees, which can be used to solve a variety of predictive modeling tasks, including structured output prediction.
Combining them into ensembles of PCTs yields state-of-the-art performance. Furthermore, the ensembles of PCTs can be interpreted by calculating feature importance scores from the learned models.
However, their learning time scales poorly with the dimensionality of the output space. 
This is often problematic, especially in (hierarchical) multi-label classification, where the output can consist of hundreds or thousands of potential labels.
Also, learning of PCTs can not exploit the sparsity of data to improve computational efficiency.
Data sparsity is common in both input (molecular fingerprints, bag of words representations) and output spaces (in multi-label classification, examples are often labeled with only a handful of labels out of a much larger set of potential labels).
In this paper, we propose oblique predictive clustering trees, capable of addressing these limitations.
We design and implement two methods for learning oblique splits that contain linear combinations of features in the tests, hence a split corresponds to an arbitrary hyperplane in the input space. The resulting oblique trees are efficient for high dimensional data and are capable of exploiting sparse data.
We experimentally evaluate the proposed methods on 60 benchmark datasets for 6 predictive modeling tasks: binary classification, multi-class classification, multi-label classification, hierarchical multi-label classification, single-target regression, and multi-target regression.
The results of the experiments show that oblique predictive clustering trees achieve performance on par with state-of-the-art methods and are orders of magnitude faster than standard predictive clustering trees.
We also show that meaningful feature importance scores can be extracted from the models learned with the proposed methods.
\end{abstract}


\end{frontmatter}

%\linenumbers

\section{Introduction}
In predictive modeling, a set of learning examples is used to induce a model that can be used to make accurate predictions for unseen examples. 
The examples are described with features and are associated with a target variable. 
The model uses the features to predict the value of the target variable. 
In most common tasks, there is only one target variable.
If it is numeric, the task is called regression; if the target is discrete, the task is called classification.

However, many real-life problems can be more naturally represented with more complex targets composed of multiple variables that can have additional structure or dependencies among them.
Such problems are encountered in a variety of disciplines: life sciences (e.g., gene and protein function prediction), environmental sciences (e.g., soil functions, habitat modeling), text and image analysis (e.g., document classification, image annotation) and others.
These problems are addressed with the task of \emph{structured output prediction} (SOP). In this work, we focus on three types of SOP tasks:
\begin{itemize}
    \item \emph{Multi-target regression} (MTR) is a predictive modeling task where the target is a vector with numeric components/variables.
    \item \emph{Multi-label classification} (MLC) is a generalization of the standard classification where each example can be a member of multiple classes: the targets are subsets of a finite set of possible labels.
    \item \emph{Hierarchical multi-label classification} (HMLC) further extends the task of MLC by additionally considering a partial order on the set of possible labels, i.e., some labels are specialized cases of other labels. The partial order organizes the labels into a hierarchy that can be represented with a directed acyclic graph.
\end{itemize}
%In multi-target regression (MTR) the target variables are vectors with numeric components.
%Multi-label classification (MLC) is a generalization of the standard classification where each example can be a member of multiple classes: the targets are subsets of a finite set of possible labels.
%If we additionally have a partial ordering on the set of possible labels (i.e., some labels are specialized cases of other labels), the task is known as hierarchical multi-label classification (HMLC).
%The partial ordering organizes the labels into a hierarchy, which can be represented with a directed acyclic graph.

Predictive clustering trees are a variant of decision trees that have been successfully applied to various predictive modeling tasks, including structured output prediction and semi-supervised learning.
However, they scale poorly to problems with many target variables and cannot take advantage of sparsity in the data.
Both of these properties are quite common, especially in structured output prediction problems.
For example, in (H)MLC problems there are often hundreds or thousands of possible labels (many target variables).
Additionally, each example is typically labeled with only a handful of labels, which leads to sparse target matrices.
Data can also be sparse on the input side, e.g., compounds described with binary fingerprints, text described with bag of words representations, etc.

In this paper, we propose two methods for learning oblique predictive clustering trees, designed to improve the efficiency of learning on high dimensional and/or sparse data. 
The first variant of oblique predictive clustering trees is based on SVM and the second on gradient descent. 
We experimentally evaluate the proposed methods and show that they achieve state-of-the-art performance and are learned orders of magnitude faster than axis-parallel trees.
We also perform parameter sensitivity analysis and demonstrate that meaningful feature importance scores can be extracted from the learned models.

Initial experiments on classification with the gradient descent variant were presented in \cite{ismis:spyct-lncs}.
The method has since been revised and improved with better support sparse data and regularization to reduce model sizes. In addition to the initial study, this paper proposes a new method (the SVM-based) and more extensive experiments that include single-target classification and regression as well as the three above-mentioned SOP tasks (MTR, MLC, and HMLC).

The remainder of the paper is organized as follows.
In Section~\ref{sec:background}, we present the background related to this paper. 
In Section~\ref{sec:methods}, we describe and analyze our proposed methods in detail. 
Next, we present our experimental setup (Section~\ref{sec:experimental}) and results from the benchmarking experiments (Section~\ref{sec:results}).
We then illustrate the meaningfulness of the feature importances extracted from our models (Section~\ref{sec:fimpresults}) and present the results of parameter sensitivity analysis (Section~\ref{sec:parameters}).
In Section~\ref{sec:conclusion}, we conclude the paper with a summary of the main findings.

\section{Background}
\label{sec:background}
The research of tree-based predictive models was popularized in the 1980s \cite{Breiman84:CART}. 
Their use is widespread: they can be used for classification and regression tasks and can handle both numeric and nominal features.
A single tree can be inspected and its predictions interpreted easily. 
When used in ensembles \citep{Breiman96:BAG,Breiman01:RF} they can achieve state-of-the-art performance. The performance boost comes at the cost of reduced interpretability, hence, different feature ranking approaches based on trees and tree ensembles have also been developed \citep{Breiman01:RF,Petkovic19:FIMP-MTR}.

Predictive clustering trees \cite{Blockeel98:PCT,Blockeel02:PCT} generalize standard decision/regression trees by differentiating between three types of attributes: features, clustering attributes, and targets.
Features are used to divide the examples; these are the attributes encountered in the split nodes. 
Clustering attributes are used to calculate the heuristic which guides the search of the best split at a given node. 
Targets are the attributes predicted in the leaves.
The algorithm for learning PCTs follows the top-down induction algorithm described by \cite{Breiman84:CART}, and is presented in Algorithm~\ref{alg:pct}.

\begin{algorithm}[tb]
\caption{Learning a PCT: The inputs are matrices of features $X \in \mathcal{R}^{N \times D}$, targets $Y \in \mathcal{R}^{N \times T}$ and clustering attributes $Z \in \mathcal{R}^{N \times K}$.}

\begin{algorithmic}[1]
  \Procedure{grow\_tree}{X, Y, Z}
  \State test = best\_test(X, Z)
  \If{acceptable(test)}
  \State rows1, rows2 = split(X, test)
  \State left\_subtree = grow\_tree(X[rows1], Y[rows1], Z[rows1])
  \State right\_subtree = grow\_tree(X[rows2], Y[rows2], Z[rows2])
  \State {\bf return} Node(test, left\_subtree, right\_subtree)
  \Else
  \State {\bf return} Leaf(prototype(Y))
  \EndIf
  \EndProcedure

  \Procedure{best\_test}{X, Z}
  \State best = None
  \For{$d = 1, \dots, D$}
  \For{test $\in$ possible\_tests(X, d)}
  \If{score(test, X, Z) $>$ score(best, X, Z)}
  \State best = test
  \EndIf
  \EndFor
  \EndFor
  \State {\bf return} best $\frac{}{}$
  \EndProcedure

  \Procedure{score}{test, X, Z}
  \State rows1, rows2 = split(X, test)
  \State n1 = num\_rows(rows1)
  \State n2 = num\_rows(rows2)
  \State n = n1 + n2
  \State {\bf return} $n \cdot$ impurity(Z) - $n1 \cdot$ impurity(Z[rows1]) - $n2 \cdot$ impurity(Z[rows2])
  \EndProcedure
\end{algorithmic}
\label{alg:pct}
\end{algorithm}

The algorithm takes as input matrices of features ($X$), clustering attributes ($Y$), and targets ($Z$).
It then goes through all features and searches for a test that maximizes the heuristic score.
The heuristic that is used to evaluate the tests is the reduction of impurity caused by splitting the data according to a test.
It is calculated on the clustering attributes.
If no acceptable split is found (e.g., no test reduces the variance significantly, or the number of examples in a node is below a user-specified threshold), then the algorithm creates a leaf and computes the prototype of the targets of the instances that were sorted to the leaf.
The selection of the impurity and prototype functions depends on the types of clustering attributes and targets (e.g., variance and mean for regression, entropy and majority class for classification).

In theory, clustering attributes can be completely independent of the features and the targets.
In practice, the ultimate goal is to make accurate predictions for the targets, and the splitting heuristic should reflect that.
The most basic (and common) approach is to use the targets also as clustering attributes.
For example, in a classification problem, doing so makes PCTs equivalent to standard decision trees.
But the attribute differentiation gives PCTs a lot of flexibility.
They have been used for predicting various structured outputs \cite{Kocev13:SOP}.
In addition to targets, we can also include features among the clustering attributes.
This makes leaves homogeneous also in the input space, which is helpful if the targets are noisy, and can also be used for semi-supervised learning \cite{Levatic17:SSL-CL,Levatic18:SSL-MTR}.
Embeddings of the targets have also been used as clustering attributes in order to reduce the time complexity of tree learning \cite{ismis:hmlc-lncs}.

\begin{figure}[bt!]
    \centering
    \begin{tabular}{cc}
    \includegraphics[width=0.41\textwidth]{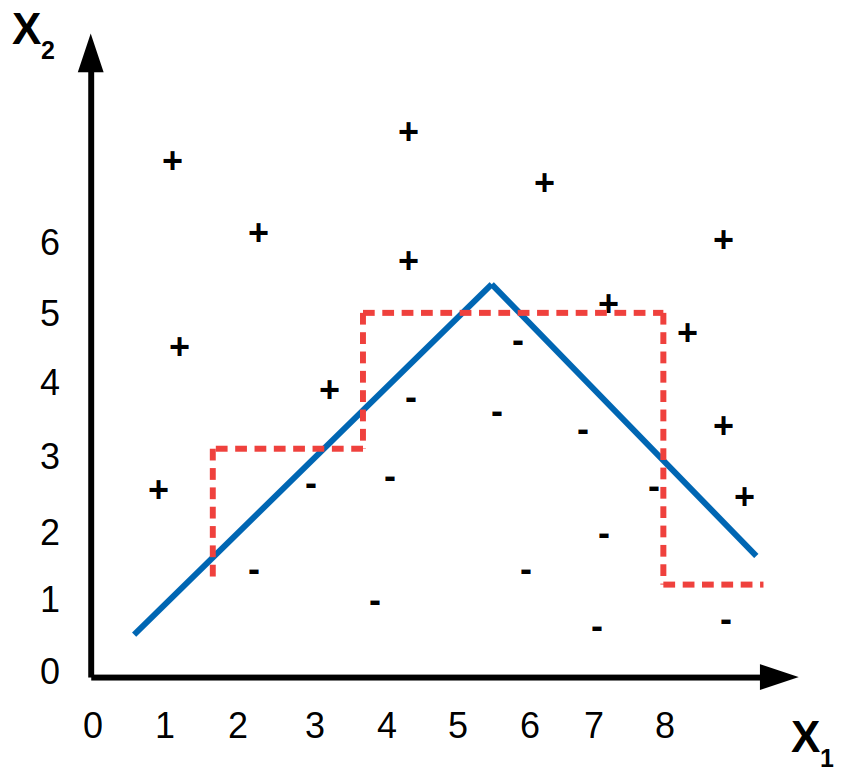} &
    \includegraphics[width=0.52\textwidth]{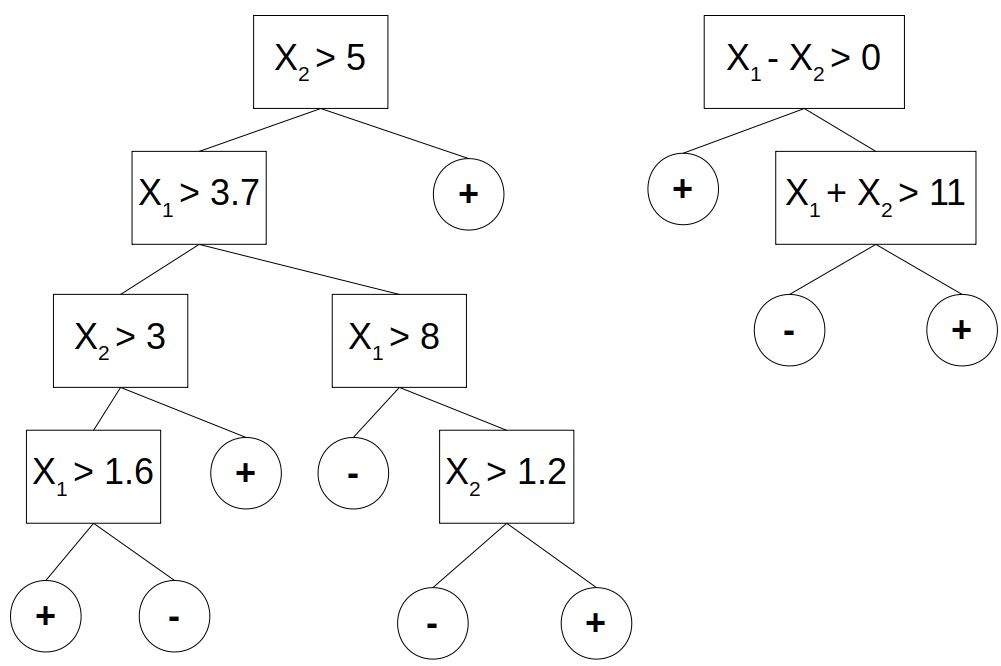} \\
    a) & b)
    \end{tabular}
    \caption{A toy dataset (a) with drawn decision boundaries learned by axis-parallel (red, dashed) and oblique (blue, solid) decision trees (b).}
    \label{fig:demo}
\end{figure}

Most of the research focus, including all of the above-mentioned work, is on \emph{axis-parallel} trees. 
The tests in axis-parallel trees only use single features and the splits are axis-parallel hyperplanes in the input space. 
Alternatively, \emph{oblique} trees (also called multivariate trees) use linear combinations of features in the tests, which allow for splits corresponding to an arbitrary hyperplane in the input space. 
They receive a lot less attention than axis-parallel trees, possibly due to the increased complexity of optimizing the split. 
Finding a hyperplane that splits a set of examples with binary labels in a way that minimizes misclassified examples on both sides of the hyperplane is an NP-hard problem \cite{Heath93:NP}.
But the increased flexibility of splits can lead to models that fit the data much better as illustrated in Figure~\ref{fig:demo}.

The initial proposals \citep{Breiman84:CART,Murthy94:OC1} for the induction of oblique trees relied on local search optimization and scale poorly to contemporary problems with thousands of examples and/or features.
Weighted Oblique Decision Trees \cite{Yang19:WODT} learn splits by optimizing weighted information entropy and are a relatively efficient method for binary and multi-class classification.
Oblique random forests \cite{Menze11:ORF} follow the standard Random Forest paradigm of learning trees on different bootstrapped samples of the training set and searching for each split in a different subset of features. 
Split learning is performed via ridge regression, which allows for efficient optimization of the hyperplanes, but their approach is limited to binary classification problems.
Additionally, FastXML \cite{Prabhu14:FastXML} and  PfastreXML \cite{Jain16:PFAST} present methods based on oblique trees for multi-label classification which work efficiently even with thousands of features and labels. 
When learning a split, these methods optimize a complex objective function that combines L1 regularization, logarithmic loss and normalized Distributed Cumulative Gain (nDCG) of examples on each side of the hyperplane. 
The criterion function optimization is performed in two steps.
First, the examples are partitioned in a way that minimizes nDCG in each partition; this partitioning is improved iteratively.
Then, a hyperplane is learned that approximates this partitioning as best as it can.
This is done by using GLMNET method to solve a logistic regression problem.

Existing oblique tree methods are specialized for specific predictive tasks and/or computationally inefficient. In this paper, we present a method that combines the flexibility of PCTs, uses oblique splits, applicable to a variety of predictive modeling tasks, and is computationally efficient.

\section{Method description}
\label{sec:methods}

In this section, we describe oblique predictive clustering trees. We also follow the top-down induction algorithm presented in Algorithm~\ref{alg:pct}, but we modify the split searching procedure (the BEST\_TEST function) to learn oblique tests. We propose two modifications of the BEST\_TEST function based on SVMs and gradient descent. 

We start with the introduction of the notation used throughout the manuscript.
Let $X \in \mathcal{R}^{N \times D}$ be the matrix containing the $D$ features of the $N$ examples in the learning set, $Y \in \mathcal{R}^{N \times T}$ be the matrix containing the $T$ targets associated with the $N$ examples in the learning set, and $Z \in \mathcal{R}^{N \times K}$ be the matrix containing the $C$ clustering attributes associated with the $N$ examples in the learning set.
This formulation encapsulates several predictive modeling tasks. 
First, single-target regression ($T=1$) and multi-target regression ($T > 1$) fit this description easily.
Second, for classification problems, the $Y$ matrix contains binary values: The value $y_{ij} = 1$ if the $i$-th example has $j$-th label, otherwise $y_{ij} = 0$. 
Third, for binary classification problems, the number of targets is $T=1$. 
Next, for multi-class classification with $k$ possible classes, the class information can be encoded via one-hot encoding ($T=k$).
Each row in $Y$ has exactly one element set to 1, and the rest are set to 0. Note that similarly as for the targets, nominal features across all tasks are supported via one-hot encoding. 
Finally, label sets in (hierarchical) multi-label classification with $k$ possible labels can be encoded with binary vectors ($T=k$), but here each row can have multiple elements set to 1.
%Finally, similarly as for the target attributes, for the descriptive attributes, the nominal attributes are supported via one-hot encoding. %To follow the predictive clustering framework, we will also treat clustering attributes separately: Let 

Below, $M_{i.}$ will refer to the $i$-th row and $M_{.i}$ to the $i$-th column of the matrix $M$.
Let $\colmean(M)$ be a vector of column means of the matrix M.
We wish to learn a vector of weights $w \in \mathcal{R}^D$ and a bias term $b \in \mathcal{R}$ that define a hyperplane, which splits the learning examples into two subsets.
We will refer to the subset where $X_{i.} \cdot w+b \geq 0$ as the positive subset, and the subset where $X_{i.} \cdot w+b < 0$ as the negative subset.
We want the examples in the same subset to have similar values of clustering attributes in $Z$.
Prior to learning each split, features and clustering attributes are standardized to mean 0 and standard deviation 1.
%Below we will present two variants for learning the split hyperplane.
After learning the split, the tree construction continues recursively on the positive and negative subsets. 
As with standard PCTs, the learning stops when the learned split is not acceptable, at which point a leaf node is made where the prototype of the targets is stored.
As prototypes, we use target means over the examples in the leaf, i.e., \colmean(Y).
This prototype can be used as-is to make predictions (regression problems, pseudo-probabilities for classification problems), or it can be used to predict the majority class, or all labels with frequencies above a specified threshold (multi-label classification).

When making a prediction for a new example with features $x \in \mathcal{R}^D$, we calculate the value $x \cdot w + b$ in the root node. 
If it is non-negative, we repeat the process in the positive subtree, otherwise in the negative subtree. 
This is done until a leaf node is reached, where a prediction is made.

\subsection{SVM-based split learning}

In the SVM-based split learning, we perform hyperplane optimization in two steps. First, we group the examples into two subsets in such a way that the similarity of clustering attributes in each subset is maximized, regardless of the values of the features. 
This is calculated using only the $Z$ matrix, while $X$ is ignored. 
For example, in binary classification the subset partitioning is clear: each subset should contain examples from one class.
For other tasks, the subsets are not obvious.
To obtain them we use the k-means clustering \cite{kmeans} to cluster rows in $Z$ into two clusters.
Let vector $c \in \{0, 1\}^N$ be the result of the clustering.
    
Next, we look for a hyperplane in the input space that approximates this partitioning. 
In essence, we convert the split hyperplane optimization problem into a binary classification problem. 
We solve the following problem:
$$\underset{w, b}{\min} \;\; ||w||_1 + C \sum_{i=1}^N max(0, 1 - c_i (X_{i.} \cdot w + b))^2,$$
where parameter $C\in \mathcal{R}$ determines the strength of regularization. 
To solve this problem we use the LIBLINEAR library \cite{liblinear}, specifically their $L1$-regularized $L2$-loss support vector classification.

\subsection{Gradient-descent-based split learning}

In the gradient-descent-based split learning, we directly search for a hyperplane split that minimizes the impurity on both sides of the hyperplane, unlike in the SVM-based split learning, where the split hyperplane is optimized indirectly, by first determining an ideal split and then trying to approximate it with a hyperplane.
First, we calculate the vector $s = \sigma(Xw+b) \in [0, 1]^N$, where the sigmoid function $\sigma$ is applied component-wise.
The vector $s$ contains values from the $[0, 1]$ interval, and we treat it as a fuzzy membership indicator.
Specifically, the value $s_i$ tells us how much the $i$-th example belongs to the positive subset, whereas the value $1-s_i$ tells us how much it belongs to the negative subset.

To measure the impurity of the positive subset, we calculate the weighted variance of each column (attribute) in $Z$, and we weigh each row (example) with its corresponding weight in $s$. To measure the impurity of the negative subset, we calculate the weighted variances with weights from $1-s$.
Weighted variance of a vector $v \in \mathcal{R}^n$ with weights $a \in \mathcal{R}^n$ is defined as
$$\var(v, a) = \frac{\sum^n_{i=1} a_i(v_i - \mean(v, a))^2}{A} = \mean(v^2, a) - \mean(v, a)^2,$$
where $A = \sum^n_{i=1} a_i$ is the sum of weights and $\mean(v, a) = \frac{1}{A}\sum^n_{i=1} a_i v_i$ is the weighted mean of $v$.

The impurity of a subset is the weighted sum of weighted variances over all the clustering attributes.
The impurity of the positive subset is
$\imp(Z, p, s) = \sum^L_{j=1} p_j \var(Z_{.j}, s)$, and similarly $\imp(Z, p, 1-s)$ is the impurity of the negative subset.
Weights $p \in \mathcal{R}^K$ enable us to give different priorities to different clustering attributes.
The final split fitness function is
$$ f(w, b) = S * \imp(Z, p, s) + (N-S) * \imp(Z, p, 1-s), $$
where $s = \sigma(Xw+b)$ and $S = \sum^N_{i=1} s_i$.
The terms $S$ and $N-S$ represent the sizes of positive and negative subsets, and are added to guide the split-search procedure towards balanced splits.
Finally, to obtain the split hyperplane, the following optimization problem is solved:
$$\underset{w, b}{\min} \;\; ||w||_\frac{1}{2} + C f(w, b),$$
where $C$ again controls the strength of regularization and $||w||_\frac{1}{2} = (\sum_{i=1}^D \sqrt{|w_i|})^2$ is the L$\frac{1}{2}$ norm.
We selected L$\frac{1}{2}$ regularization because it induces weight sparsity more aggressively than L1, which helps reducing the model size.

For examples that are not close to the hyperplane, $s_i$ is close to 0 or 1.
Weighted variances are therefore approximations of the variances of the subsets, that the hyperplane produces.
This formulation makes the objective function differentiable and enables us to use the efficient Adam \cite{adam} gradient descent optimization method. %, which is a variant of gradient descent.
The weight vector $w$ is initialized randomly, then $b$ is set such that the examples are initially split in half.

\subsection{Time complexity analysis}
\label{sec:time_complexity}
In this section, we analyze the time complexity of learning oblique predictive clustering trees and compare it to the time complexity of learning standard PCTs.
We focus on the cost of learning a split since this is the only significant difference between the two methods.

Let us start with the SVM variant. 
The first step is to perform $2$-means clustering on the clustering data $Z \in \mathcal{R}^{N \times K}$, with time complexity $O(N K I_c)$, where $I_c$ is the number of clustering iterations we perform. 
Learning the linear SVM to differentiate the clusters costs $O(N D I_o)$, where $I_o$ is the number of optimization iterations. 
The total cost is then $O(N(K I_c + D I_o))$.

For the gradient descent variant, the main operation is gradient calculation. 
The most expensive part of it is the multiplications of matrices $X$ and $Z$ with vectors, which costs $O(ND)$ and $O(NK)$, respectively. 
The total cost of learning a split is then $O(N I_o (D+K))$, where $I_o$ is again the number of optimization iterations.

The time complexity of learning a split in standard PCTs is $O(DN \log N + NDK)$ \cite{Kocev13:SOP}. 
The important difference we can notice is that standard PCTs scale with $DK$, whereas both proposed variants of oblique PCTs scale with $D+K$. 
This difference is very noticeable when solving problems with many clustering variables, e.g., (hierarchical) multi-label classification and semi-supervised learning. 
An additional benefit of our approach can be obtained when dealing with sparse data. 
If the features and/or clustering data is sparse, both variants can exploit the sparsity by performing operations with sparse matrices. 
This reduces the time complexity to $O(N(\hat{K} I_c + \hat{D} I_o))$ for the SVM variant and $O(N I_o (\hat{D}+\hat{K})$ for the gradient descent variant, where $\hat{D} \ll D$ and $\hat{K} \ll K$ are average numbers of non-zero elements in each row of matrices $X$ and $Z$.

\subsection{Ensembles of oblique predictive clustering trees}
Single decision trees are mainly used when we wish to visually inspect and interpret the model.
To achieve state-of-the-art performance, trees have to be used in ensembles.
Oblique trees are inherently more difficult to visually interpret, because of the linear combinations in the split nodes.
Also, because the splits are more complex, they can very easily overfit the data in the deeper nodes.
For these reasons, we believe the most natural use of oblique PCTs is by combining them in ensembles.

To construct bagging ensembles \cite{Breiman96:BAG}, we simply construct each oblique PCT on a different bootstrapped sample of the learning set.
When making predictions, we average the prototypes predicted by each tree in the ensemble to get the final prediction.
We can also build random forest ensembles \cite{Breiman01:RF} of oblique PCTs.
In addition to bootstrapping, each split hyperplane is only learned on a random subset of features.
After the hyperplane weights are learned, the $w_i$ for features that were not in the selected subset are set to $0$.

\subsection{Feature importance}
\label{sec:fimp}
Even though ensembles of oblique PCTs are hard to interpret directly, we can still gain insight into how the models make their decision by calculating feature importances, much like it is done with ensembles of axis-parallel trees.
The feature importance scores of a single oblique PCT are calculated as follows:
$$ imp(T) = \sum_{s \in T} \frac{s_n}{N} \frac{s_w}{\Vert s_w \Vert_1}, $$
where $s$ iterates over split nodes in tree $T$, $s_w$ is the weight vector defining the split hyperplane, $s_n$ is the number of learning examples that were present in the node and $N$ is the total number of learning examples. 
The contributions of each node to the final feature importance scores are weighted according to the number of examples that were used to learn the split.
This puts more emphasis on weights higher in the tree, which affect more examples.
To get feature importance scores of an ensemble, we simply average feature importances of individual trees in the ensemble.

\subsection{Implementation}
We implemented the proposed oblique predictive clustering trees in a python package \emph{spyct}. 
It is freely licensed and available for use and download at \url{https://gitlab.com/TStepi/spyct}.
The package includes both SVM and gradient descent variant, single trees, bagging, and random forest ensembles.
In ensembles, trees can be built in parallel.
We also provide a number of pre-pruning options: maximum tree depth, minimum number of examples required to perform a split, and the minimum reduction of impurity required for a split to be accepted.
By default, tree depth is not limited, a split is always attempted if more than 1 example is still present in a node, and it is accepted if the impurity is reduced by at least 5\% in at least one of the subsets.
Additionally, the splitting stops if one of the subsets is empty (the hyperplane does not split the data).
We can control the maximum number of clustering iterations (SVM variant, 10 by default), learning rate (gradient variant, default 0.1), and the maximum number of hyperplane optimization iterations (both variants, default 100).
We also make use of early stopping in both clustering and hyperplane optimization, if the process converges.
Strength of regularization (both variants, $C=10$ by default) and other parameters of the Adam optimizer (gradient variant, default values from PyTorch \footnote{\url{https://pytorch.org/docs/stable/optim.html}} library) are also configurable.

\section{Experimental setting}
\label{sec:experimental}
In this section, we present a comprehensive experimental study designed to evaluate the proposed oblique PCTs and compare them to standard PCTs and other baseline methods.
We evaluated the methods on benchmark datasets from the classification spectrum: binary (BIN), multi-class (MCC), multi-label (MLC), and hierarchical multi-label classification (HMLC), as well as from the regression spectrum: single-target (STR) and multi-target regression (MTR). We first present the competing methods, and then the benchmark data and evaluation strategy.

\subsection{Competing methods}
The main baselines for comparison are standard PCTs, which can be used for all 6 predictive modeling tasks considered. 
We trained single trees, bagging ensembles of PCTs, and random forests of PCTs.
Next, we compared to existing oblique tree methods in single tree settings, as they were proposed by the authors.
This includes CART-LC \cite{Breiman84:CART}, OC1 \cite{Murthy94:OC1} and WODT \cite{Yang19:WODT}.
Both CART-LC and OC1 are only applicable to BIN problems, whereas WODT can also be used for MCC.

For ensembles of oblique trees, we considered oblique random forests (ORF) \cite{Menze11:ORF} and the FastXML method \cite{Prabhu14:FastXML} as competing methods. 
The ORF method is only applicable to BIN problems.
The \textsc{FastXML} method is specialized for the MLC task and optimized to work with sparse data.
We will also use it as a baseline for the HMLC datasets, by discarding the hierarchy and treating the problem as a flat MLC task.

Additionally, the comparison includes \textsc{LightGBM} gradient boosted tree ensembles \cite{lightGBM}. 
They can be used for BIN, MCC, and STR tasks. 
Finally, to include a non-tree-based baseline, we used linear SVMs. They can be directly applied to BIN and STR tasks. 
We will also use them as baselines for other tasks, by learning a separate SVM for each label/target (one-vs-all approach in MCC, binary relevance in MLC and HMLC, local approach to MTR).
We decided to use the linear kernel to keep the learning time manageable and because it is closest to the nature of the splits used in the oblique trees.

Table~\ref{tab:baselines} sums up the baseline methods and provides links to the implementations we used in the experiments. 
We will refer to the proposed methods as \textsc{spyct-svm} and \textsc{spyct-grad}, for the SVM and gradient descent variant, respectively. 
For standard and oblique PCTs, we will prepend \textsc{BAG} or \textsc{RF} to mark bagging and random forest ensembles, respectively (e.g., \textsc{BAG-spyct-grad} denotes bagging ensembles of gradient descent variant oblique PCTs).

\begin{table}[bt!]
    \centering
    \begin{tabular}{lrr}
    Method & Tasks & Implementation \\ \hline
    \textsc{spyct-svm} & all & \footnotesize{\url{gitlab.com/TStepi/spyct}}\\
    \textsc{spyct-grad} & all & \footnotesize{\url{gitlab.com/TStepi/spyct}}\\
    PCT & all & \footnotesize{\url{http://source.ijs.si/ktclus/clus-public/}}\\
    SVM & all & \footnotesize{\url{https://scikit-learn.org/stable/}}\\
    CART-LC & BIN & \footnotesize{\url{github.com/AndriyMulyar/sklearn-oblique-tree}}\\
    OC1 & BIN & \footnotesize{\url{github.com/AndriyMulyar/sklearn-oblique-tree}} \\
    WODT & BIN, MCC & \footnotesize{\url{www.lamda.nju.edu.cn/yangbb}}\\
    ORF & BIN & \footnotesize{\url{rdrr.io/cran/obliqueRF/man/obliqueRF.html}}\\
    \textsc{FastXML} & MLC, HMLC & \footnotesize{\url{manikvarma.org/code/FastXML/download.html}}\\
    \textsc{LightGBM} & BIN, MCC, STR & \footnotesize{\url{lightgbm.readthedocs.io}}\\
    \end{tabular}
    \caption{Methods included in the benchmarking comparison.}
    \label{tab:baselines}
\end{table}

\subsection{Evaluation}
    
\begin{table}[bt!]
    \centering
    \begin{tabular}{l r r}
    Dataset & N & D \\ \hline
    \emph{ailerons} \cite{openml}        & 13750   & 40   \\
    \emph{cpmp-2015} \cite{openml}       & 2108    & 23   \\
    \emph{cpu\_small} \cite{openml}      & 8192    & 12   \\
    \emph{elevators} \cite{openml}       & 16599   & 19   \\
    \emph{house\_8L} \cite{openml}       & 22784   & 8    \\
    \end{tabular} \; \;
    \begin{tabular}{l r r}
    Dataset & N & D \\ \hline
    \emph{puma8NH} \cite{openml}         & 8192    & 8    \\
    \emph{qsar-234} \cite{openml}        & 2145    & 1024 \\
    \emph{satellite\_image} \cite{openml} & 6435    & 36  \\
    \emph{space\_ga} \cite{openml}       & 3107    & 6    \\
    \emph{triazines} \cite{openml}       & 186     & 60   \\
    \end{tabular}
    \caption{Properties of the benchmark STR datasets. Columns show the number of examples (N) and the number features (D).}
    \label{tab:datasets_str}
\end{table}

\begin{table}[bt!]
    \centering
    \begin{tabular}{l r r r}
    Dataset & N & D & T  \\ \hline
    \emph{atp1d} \cite{mulan}   & 337     & 411   & 6      \\
    \emph{edm} \cite{mulan}     & 154     & 16    & 2      \\ 
    \emph{enb} \cite{mulan}     & 768     & 8     & 2      \\
    \emph{jura} \cite{mulan}    & 359     & 15    & 3      \\
    \emph{oes10} \cite{mulan}   & 403     & 298   & 16     \\
    \end{tabular} \; \;
    \begin{tabular}{l r r r}
    Dataset & N & D & T \\ \hline
    \emph{oes97} \cite{mulan}   & 334     & 263   & 16     \\
    \emph{rf1} \cite{mulan}     & 9125    & 64    & 8      \\
    \emph{rf2} \cite{mulan}     & 9125    & 576   & 8      \\
    \emph{scm1d} \cite{mulan}   & 9803    & 280   & 16     \\
    \emph{slump} \cite{mulan}   & 103     & 7     & 3      \\
    \end{tabular}
    \caption{Properties of the benchmark MTR datasets. Columns show the number of examples (N), the number of features (D) and the number of targets (T).}
    \label{tab:datasets_mtr}
\end{table}
   
\begin{table}[bt!]
    \centering
    \begin{tabular}{l r r}
    Dataset & N & D \\ \hline
    \emph{banknote} \cite{openml}        & 1372 & 4      \\
    \emph{bioresponse} \cite{openml}     & 3751 & 1776   \\
    \emph{credit-approval} \cite{openml} & 690  & 15     \\
    \emph{credit-g} \cite{openml}        & 1000 & 20     \\
    \emph{diabetes} \cite{openml}        & 768  & 8      \\
    \end{tabular} \; \;
    \begin{tabular}{l r r}
    Dataset & N & D \\ \hline
    \emph{musk} \cite{openml}            & 6598 & 166    \\
    \emph{OVA\_Breast} \cite{openml}     & 1545 & 10935  \\
    \emph{OVA\_Lung} \cite{openml}       & 1545 & 10935  \\
    \emph{spambase} \cite{openml}        & 4601 & 57     \\
    \emph{speeddating} \cite{openml}     & 8378 & 120    \\
    \end{tabular}
    \caption{Properties of the benchmark BIN datasets. Columns show the number of examples (N) and the number of features (D).}
    \label{tab:datasets_bin}
\end{table}
    
\begin{table}[bt!]
    \centering
    \begin{tabular}{l r r r}
    Dataset & N & D & T \\ \hline
    \emph{amazon-reviews} \cite{openml}  & 1500    & 10000 & 50  \\
    \emph{balance} \cite{openml}         & 625     & 4     & 3   \\
    \emph{diabetes130us} \cite{openml}   & 101766  & 47    & 3   \\
    \emph{gas-drift} \cite{openml}       & 13910   & 128   & 6   \\
    \emph{hepatitisC} \cite{openml}      & 283     & 54621 & 3   \\
    \end{tabular} \; \;
    \begin{tabular}{l r r r}
    Dataset & N & D & T \\ \hline
    \emph{isolet} \cite{openml}          & 7797    & 617   & 26  \\
    \emph{mfeat-pixel} \cite{openml}     & 2000    & 240   & 10  \\
    \emph{micro-mass} \cite{openml}      & 571     & 1300  & 20  \\
    \emph{vehicle} \cite{openml}         & 846     & 18    & 4  \\
    \emph{wine-white} \cite{openml}      & 4898    & 11    & 7  \\
    \end{tabular}
    \caption{Properties of the benchmark MCC datasets. Columns show the number of examples (N), the number of features (D) and the number of classes (T).}
    \label{tab:datasets_mcc}
\end{table}

\begin{table}[bt!]
    \centering
    \begin{tabular}{l r r r r}
    Dataset & N & D & T & Sparse \\ \hline
    \emph{bibtex} \cite{mulan}      & 7395    & 1836  & 159   & D, T \\
    \emph{bookmarks} \cite{mulan}   & 87856   & 2150  & 208   & D, T \\
    \emph{CAL500} \cite{mulan}      & 502     & 68    & 174   & T \\
    \emph{corel5k} \cite{mulan}     & 5000    & 499   & 374   & D, T \\
    \emph{emotions} \cite{mulan}    & 593     & 72    & 6     & \\
    \emph{eurlex-eurovoc} \cite{mulan} & 19348 & 5000 & 3993  & D, T \\
    \emph{flags} \cite{mulan}       & 194     & 19    & 7     & \\
    \emph{rcv1subset1} \cite{mulan} & 6000    & 47236 & 101   & D, T \\
    \emph{scene} \cite{mulan}       & 2407    & 294   & 6     & \\
    \emph{tmc2007} \cite{mulan}     & 28596   & 49060 & 22    & D \\
    \end{tabular}
    \caption{Properties of the benchmark MLC datasets. Columns show the number of examples (N), the number of features (D), the number of targets/labels (T), and whether the input or output space is sparse.}
    \label{tab:datasets_mlc}
\end{table}

\begin{table}[bt!]
    \centering
    \begin{tabular}{l r r r r}
    Dataset & N & D & T & Sparse \\ \hline
    \emph{enron} \cite{kocev_phd}       & 1648   & 1001  & 56    & D, T \\
    \emph{imclef07d} \cite{kocev_phd}   & 11006  & 80    & 46    & T \\
    \emph{reuters} \cite{kocev_phd}     & 6000   & 47235 & 102   & D, T \\
    \emph{wipo} \cite{kocev_phd}        & 1710   & 74435 & 188   & D, T \\
    \emph{yeast\_GO} \cite{schietgat}   & 3465   & 5930  & 133   & D, T \\
    \emph{yeast\_spo\_FUN} \cite{schietgat}     & 3711   & 80    & 594   & T \\
    \emph{yeast\_expr\_FUN} \cite{schietgat}    & 3788   & 551   & 594   & T \\
    \emph{ara\_exprindiv\_FUN} \cite{schietgat} & 3496   & 1251  & 261   & T \\
    \emph{yeast\_gasch1\_FUN} \cite{schietgat}  & 3773   & 173   & 594   & T \\
    \emph{ara\_interpro\_GO} \cite{schietgat}   & 11763  & 2815  & 630   & D, T \\
    \end{tabular}
    \caption{Properties of the benchmark HMLC datasets. Columns show the number of examples (N), the number of features (D), the number of targets/labels (T), and whether the input or output space is sparse.}
    \label{tab:datasets_hmlc}
\end{table}
 
We evaluated the methods on 60 benchmarking datasets, 10 for each of the 6 tasks. 
Their details are presented in Tables~\ref{tab:datasets_str}-\ref{tab:datasets_hmlc}.
If the input or output data matrix had fewer than 10\% of nonzero values, it was represented in a sparse format (marked in the tables, where applicable).
To measure predictive performance of the methods on STR and MTR datasets, we used the \emph{coefficient of determination} as performance measure
$$R^2(y, \hat{y}) = 1 - \frac{\sum_i (y_i - \hat{y}_i)^2}{\sum_i (y_i - \bar{y})^2},$$
where $y$ is the vector of true target values, $\bar{y}$ is their mean, and $\hat{y}$ is the vector of predicted values.
For MTR problems, the mean of $R^2$ scores per target was calculated. 
For BIN and MCC tasks, we used \emph{F1 score}, macro averaged in the MCC case. 

Methods solving MLC and HMLC tasks typically return a score for each label and each example, higher score meaning that example is more likely to have that label.
Let $y \in \{0, 1\}^{n \times l}$ be the matrix of label indicators and $\hat{y} \in \mathcal{R}^{n \times l}$ the matrix of label scores returned by a method.
We measured the performance of methods with weighted label ranking average precision
$$LRAP(y, \hat{y}) = \frac{1}{n} \sum_{i=0}^{n-1} \sum_{j: y_{ij}=1} \frac{w_j}{W_i} \frac{L_{ij}}{R_{ij}},$$
where $L_{ij} = |\{ k: y_{ik} = 1 \wedge \hat{y}_{ik} \geq \hat{y}_{ij} \}|$ is the number of real labels assigned to example $i$ that the method ranked higher than label $j$, 
$R_{ij} = |\{k: \hat{y}_{ik} \geq \hat{y}_{ij} \}|$ is the number of all labels ranked higher than label $j$,
$w_j$ is the weight we put to label $j$ and $W_i$ is the sum of weights of all labels assigned to example $i$.
For MLC tasks we put equal weights to all labels, whereas for HMLC task we weighted each label with $0.75^d$, where $d$ is the depth of the label in the hierarchy \cite{Kocev13:SOP}.

For all three measures used (R2, F1, LRAP), a higher value indicates better performance, with a value of 1 indicating the best possible performance. 
We estimated the predictive performance using 10-fold cross-validation on each dataset. 
In addition to performance measures, we also recorded the learning time of each method.
All experiments were performed on the same computer and the methods were allowed to use up to 10 processor cores.
We set the number of trees in the ensembles to 50. 
Random forest ensembles (RF-PCT, RF-spyct-svm, RF-spyct-grad, ORF) used $\sqrt{D}$ features for each split, where $D$ is the number of features.
For SVM and LightGBM methods, parameter tuning is advised. 
In experiments with these methods, we set aside 20\% of the training set for each fold to tune the parameters.
For SVM we selected the regularization parameter $C$ from values $\{ 0.1, 1, 10, 100 ,1000 \}$. 
For LightGBM we selected the maximum number of leaves in the trees among values $\{20, 50, 100\}$ and the minimum number of samples to perform a split from values $\{1, 10, 100, 1000\}$. 
The ORF method includes internal optimization of its regularization parameter, which we left at its default setting.
In the HMLC experiments, standard and oblique PCTs also received the same label weights used to calculate LRAP.

\section{Results and discussion}
\label{sec:results}
In this section, we present the results of our extensive benchmarking experiments. 
We first discuss the predictive performance of the proposed oblique PCTs in an ensemble setting.
Next, we analyze the learning times, focusing on large and sparse datasets.
Finally, we present a comparison of the proposed methods to other single-tree methods.

\subsection{Predictive performance}

\begin{figure}[bt!]
    \centering
    \begin{tabular}{cc}
    \includegraphics[width=0.45\textwidth]{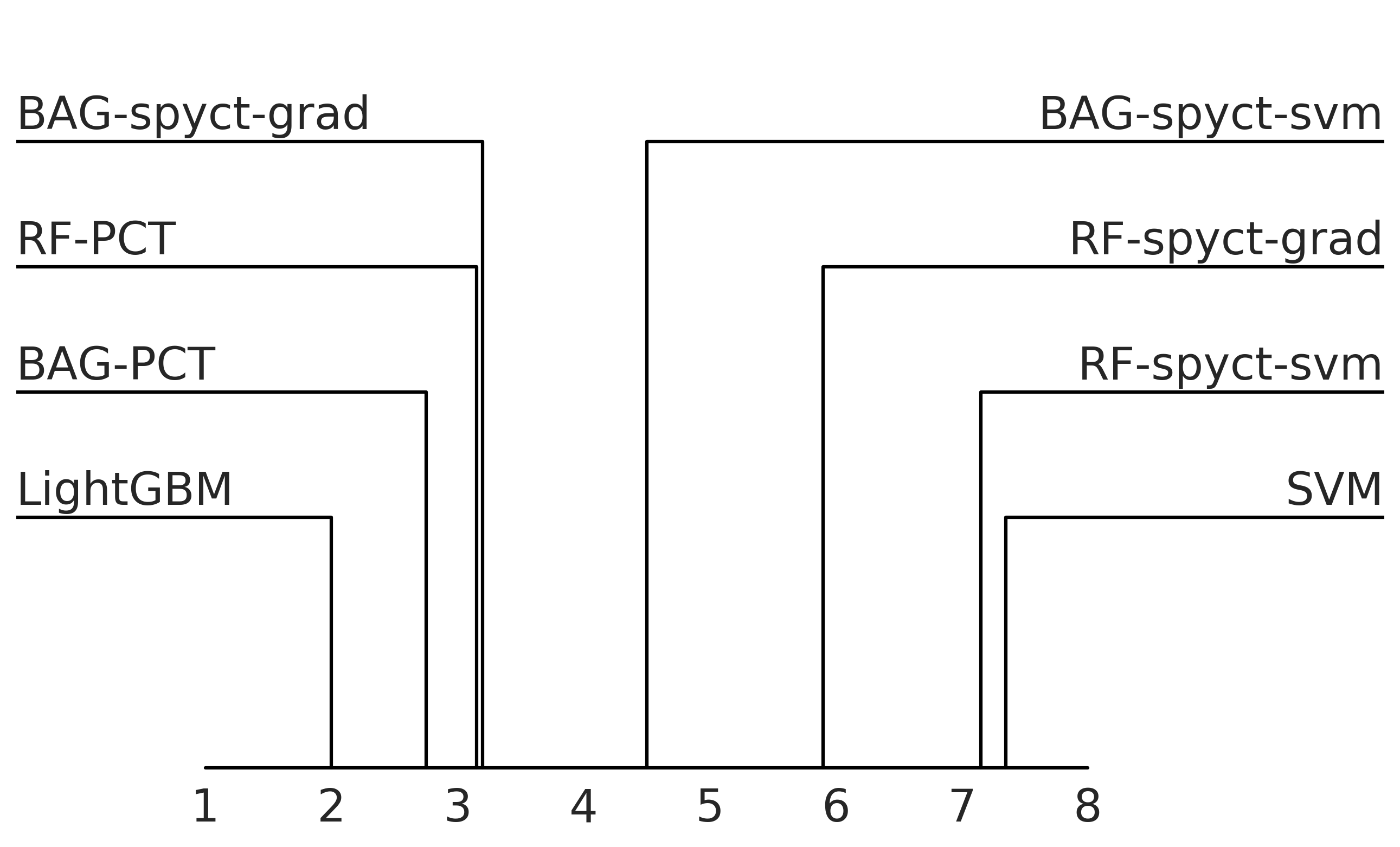} &
    \includegraphics[width=0.45\textwidth]{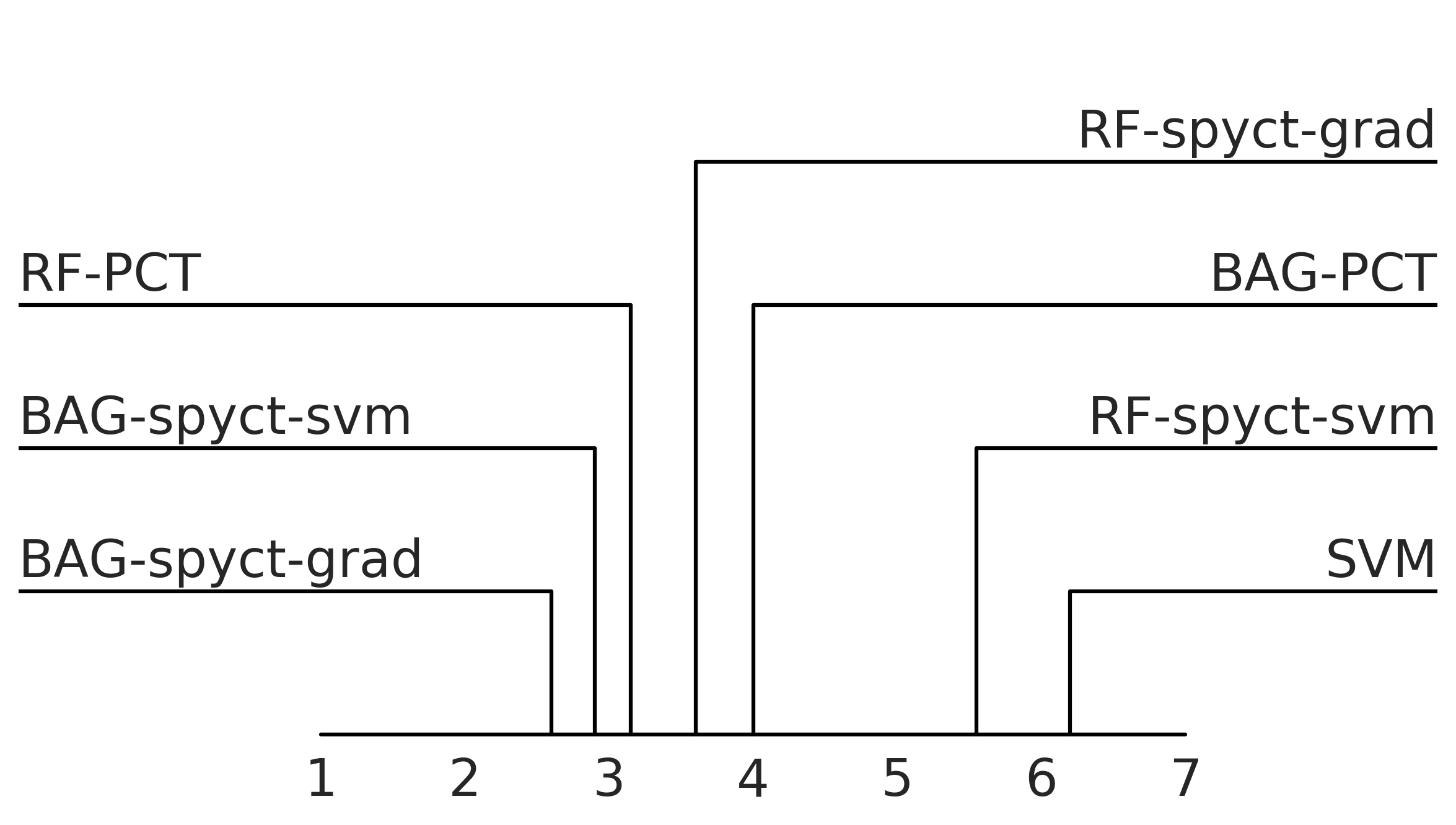} \\
    a) STR & b) MTR \\ & \\
    \includegraphics[width=0.45\textwidth]{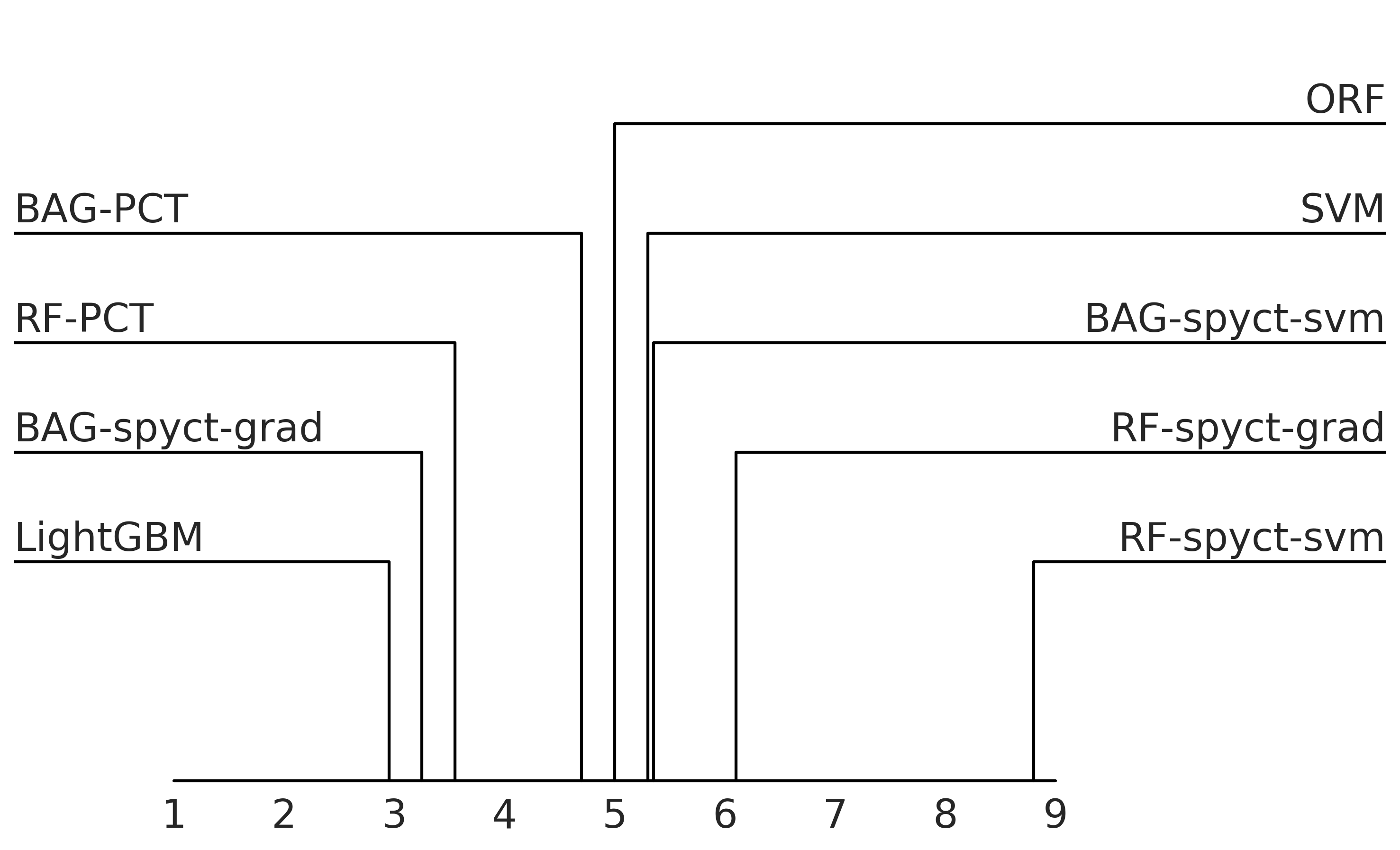} &
    \includegraphics[width=0.45\textwidth]{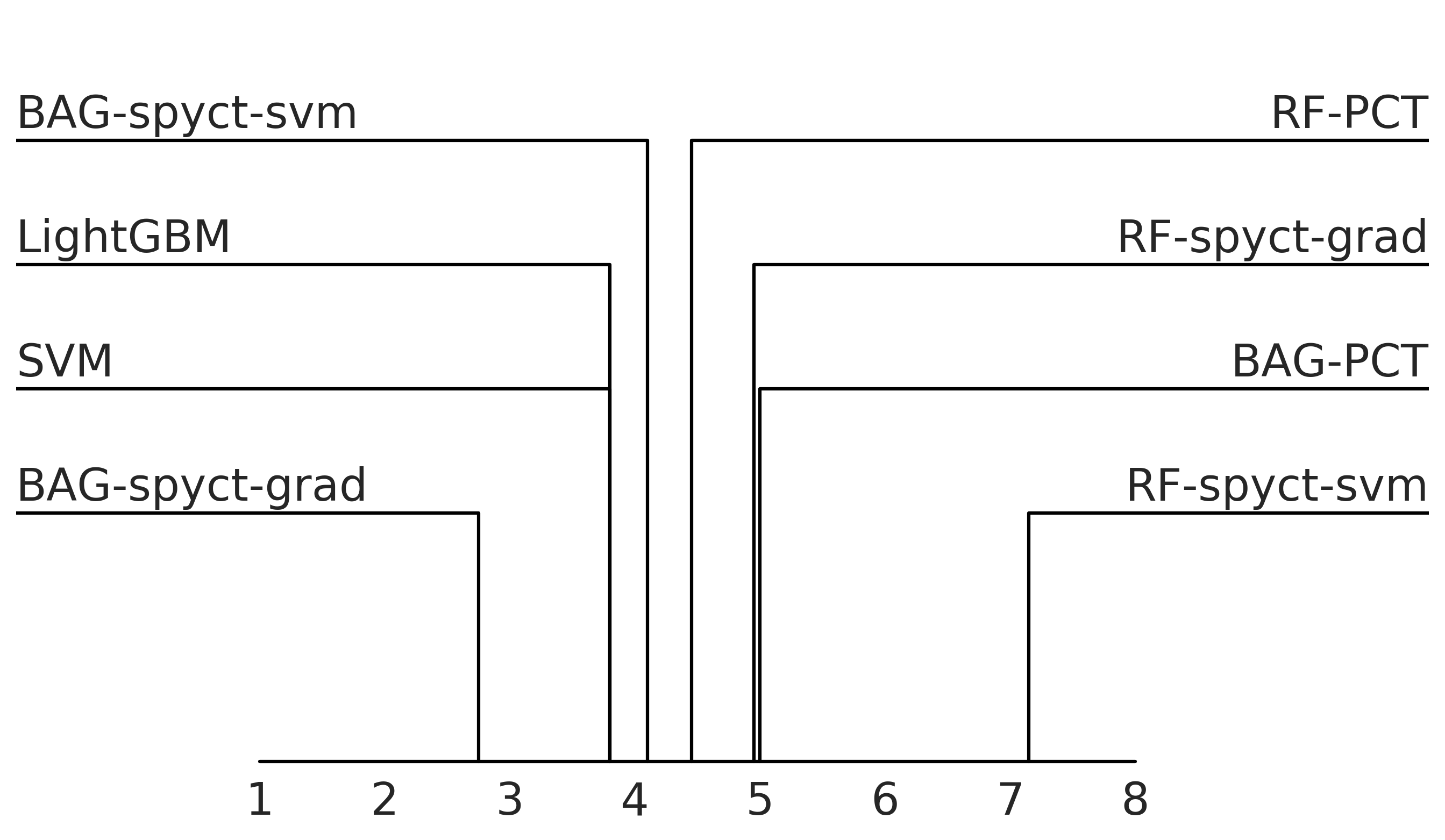} \\
    c) BIN & d) MCC \\ & \\
    \includegraphics[width=0.45\textwidth]{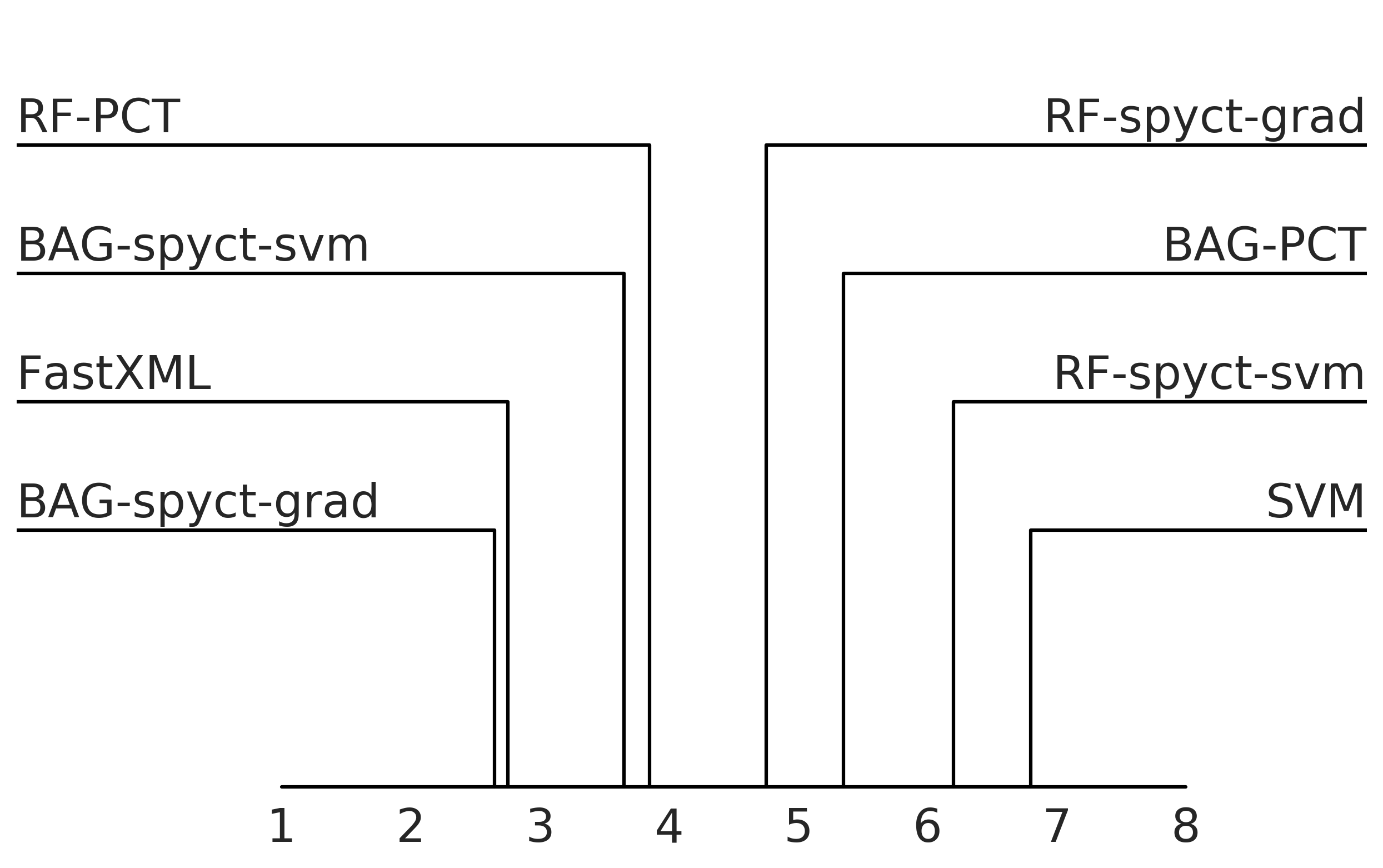} &
    \includegraphics[width=0.45\textwidth]{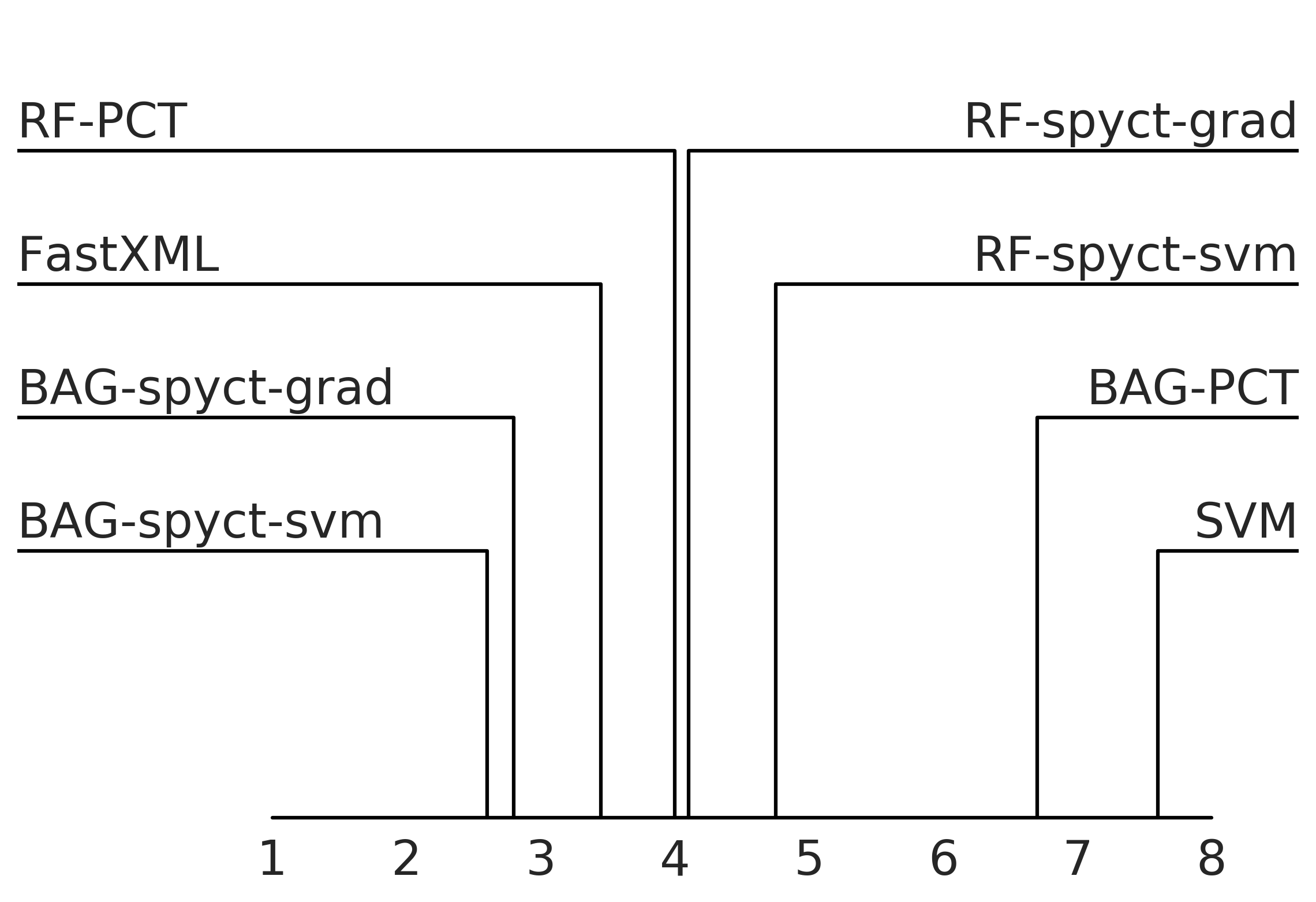} \\
    e) MLC & f) HMLC
    \end{tabular}
    \caption{Average ranks of predictive performances obtained by methods on single-target regression (a), multi-target regression (b), binary classification (c), multi-class classification (d), multi-label classification (e) and hierarchical multi-label classification (f) benchmark datasets.}
    \label{fig:ensemble_diagrams}
\end{figure}

To keep the paper concise, we do not present the large number of raw performance results of our experiments here (they are available in the Appendix), but instead focus on the aggregated results presenting the overall findings.
Figure~\ref{fig:ensemble_diagrams} presents the average ranks of the methods based on predictive performance on the benchmark datasets, grouped by predictive modeling tasks.
If a method did not finish the evaluation in 3 days, it was assigned the lowest rank.
This only occurred for the MLC task, where BAG-PCT did not finish on \emph{eurlex-eurovoc} and \emph{tmc2007} datasets, and RF-PCT did not finish on \emph{eurlex-eurovoc} dataset.

In a nutshell, the results show that in terms of predictive performance, the proposed bagging ensembles (\textsc{BAG-spyct-svm} and \textsc{BAG-spyct-grad}) are on par with current state-of-the-art methods.
Furthermore, \textsc{BAG-spyct-grad} achieves the best rank on 3 tasks (MTR, MCC, and MLC), and is close to the top on the other 3 tasks as well (close second on BIN and HMLC, where on the latter best performing is \textsc{BAG-spyct-svm}).

While the bagging ensembles of oblique PCTs have premium predictive performance, the proposed random forest ensembles of oblique PCTs (\textsc{RF-spyct-svm} and \textsc{RF-spyct-grad}) did not perform well. 
Especially \textsc{RF-spyct-svm} struggled to learn on several datasets.
We believe this is the result of sparse input features (exacerbated by the one-hot encoding required for nominal features) combined with the stopping criterion used.
A poor feature subset selection consisting of irrelevant features and features with most or all values of 0, makes split optimization difficult.
If the learned split is not useful, splitting is immediately stopped on the current branch, hence the trees become heavily over-pruned.
It appears that random forest ensembles of oblique PCTs require a deeper redesign of the learning algorithm and straight-forward subspacing is not effective.

\subsection{Learning time}

Experimental comparison of the learning times of different methods is a challenging task and difficult to do completely fairly.
The implementations are in different programming languages (python, java, R, C++) and they are not equally optimized for efficient execution.
Hence, we focus on the largest datasets, where time efficiency is more pronounced, and where the method's time complexity is more likely to dominate any implementation details.

\begin{table}
    \centering
    \begin{tabular}{l r R{1.175cm} R{1.175cm} R{1.175cm} R{1.175cm} R{1.175cm}}
    Dataset & Task & \textsc{BAG-spyct-svm} & \textsc{BAG-spyct-grad} & BAG-PCT & \textsc{FastXML} & \textsc{LightGBM}\\ \hline
    \emph{diabetes130us} & MCC & 95.7 & 65.4 & 749.0 & NA & 11.2 \\
    \emph{hepatitisC} & MCC & 5.4 & 1.8 & 12.4 & NA & 55.4 \\
    \emph{bookmarks} & MLC & 28.4 & 44.4 & 1747.7 & 16.8 & NA \\
    \emph{eurlex-eurovoc} & MLC & 200.8 & 8.0 & DNF & 8.1 & NA \\
    \emph{rcv1subset1} & MLC & 6.0 & 18.2 & 987.2 & 0.4 & NA \\
    \emph{tmc2007} & MLC & 22.5 & 129.2 & DNF & 4.8 & NA \\
    \emph{reuters} & HMLC & 4.0 & 11.8 & 883.2 & 0.4 & NA \\
    \emph{yeast\_expr\_FUN} & HMLC & 17.2 & 1.2 & 517.1 & 3.0 & NA 
    \end{tabular}
    \caption{Learning times of bagging ensembles on large datasets in minutes. NA means the method is not applicable to the task, DNF means the method did not finish in 3 days (4320 minutes). For dataset properties see Tables~\ref{tab:datasets_mcc}-\ref{tab:datasets_hmlc}.}
    \label{tab:times}
\end{table}

Table~\ref{tab:times} presents the learning times on selected datasets with large numbers of examples, features, and/or targets.
It is apparent, that both proposed variants \textsc{spyct-svm} and \textsc{spyct-grad} are computationally much more efficient than standard PCTs. 
There are several reasons for the improvement of the learning time. To begin with, the main advantage comes from better scaling with the number of targets (as theoretically analyzed in Section~\ref{sec:time_complexity}), which is evident in the large differences on MLC and HMLC datasets.
Next, another advantage is the exploitation of the sparse representation of the data.
For example, if dense matrices are used instead of sparse matrices on the \emph{reuters} dataset the learning time is dramatically increased: for \textsc{BAG-spyct-svm} it increases from 4 minutes to 247.7 minutes, and for \textsc{BAG-spyct-grad} from 11.8 minutes to 54 minutes.
Although learning from dense matrices is much slower than learning on sparse matrices, it is still faster than \textsc{BAG-PCT}.
Furthermore, another source of learning time improvements is that with oblique splits, models can be much smaller (in terms of numbers of split nodes) compared to axis-parallel trees, because the splits are more expressive. 
An illustrative example of this is the \emph{diabetes130us} dataset that does not even have multiple targets or sparse features:
The \textsc{BAG-spyct-svm} model consists of only 87 nodes and the \textsc{BAG-spyct-grad} of 2345 nodes, while the standard \textsc{BAG-PCT} model consists of 1803069 nodes.
Lastly, we noticed faster learning on datasets with a large number of features, even when they are not sparse (e.g., the learning times of the \emph{hepatitisC} dataset).
We believe this is because the matrix operations are very well optimized in modern CPUs, which gives an advantage to the proposed optimization of oblique splits, compared to exhaustive search over the features in axis-parallel trees.

We can also note that \textsc{LightGBM} is very efficient on datasets with many examples (\emph{diabetes130us}), but less so on datasets with many features (\emph{hepatitisC}).
Furthermore, the implementation of the \textsc{FastXML} method is extremely well optimized for (H)MLC datasets with a sparse structure. 
Therefore, even though its theoretical computational complexity is very similar to our proposed methods, it has lower learning times on such datasets.
However, when the features were not sparse, the observed learning times were in the same range as the proposed methods (e.g., for the \emph{yeast\_expr\_FUN dataset}, \textsc{FastXML} has worse learning time than \textsc{BAG-spyct-grad} and better than \textsc{BAG-spyct-svm}).

\subsection{Single tree methods}

\begin{figure}
    \centering
    \begin{tabular}{cc}
    \includegraphics[width=0.45\textwidth]{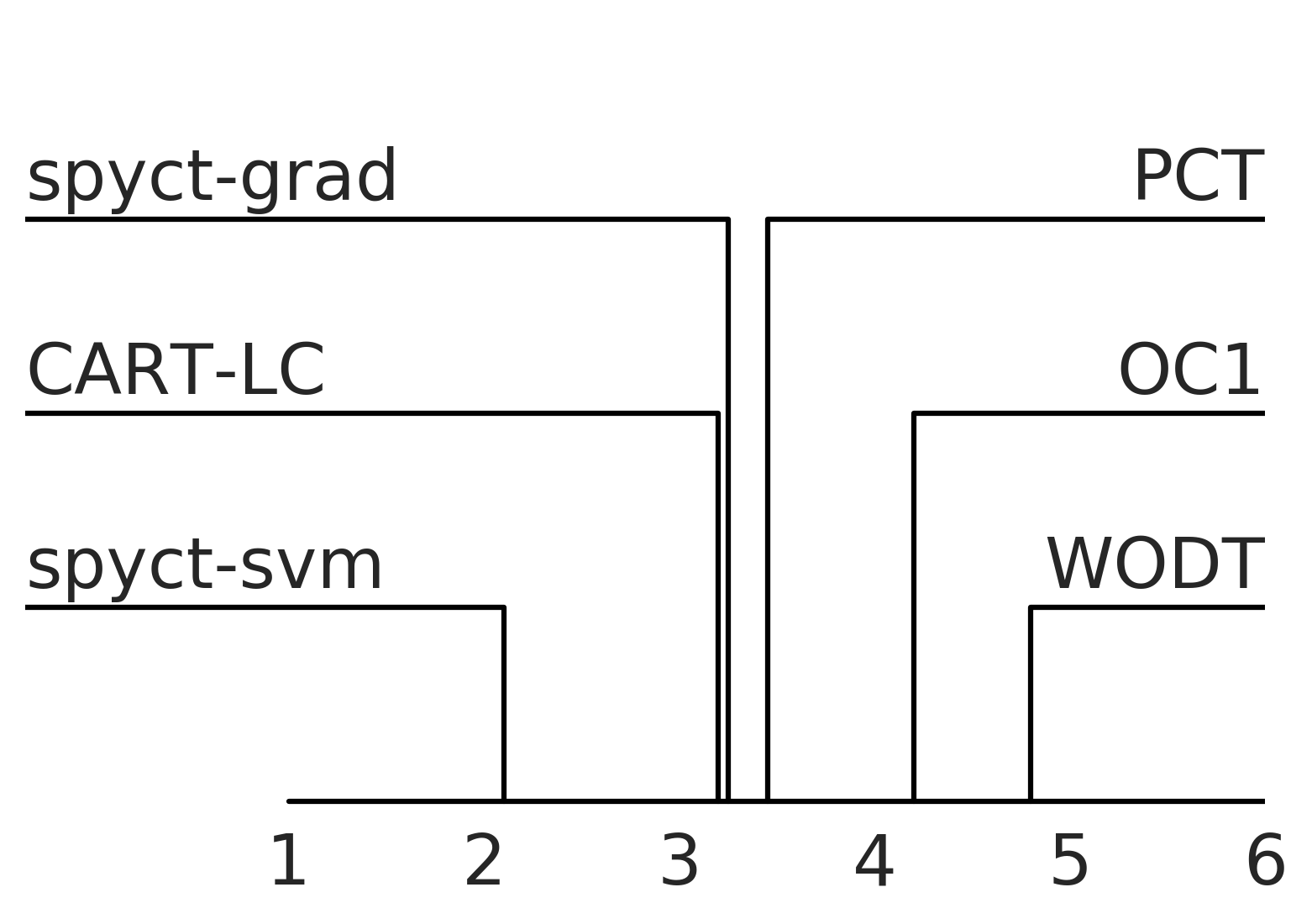} &
    \includegraphics[width=0.45\textwidth]{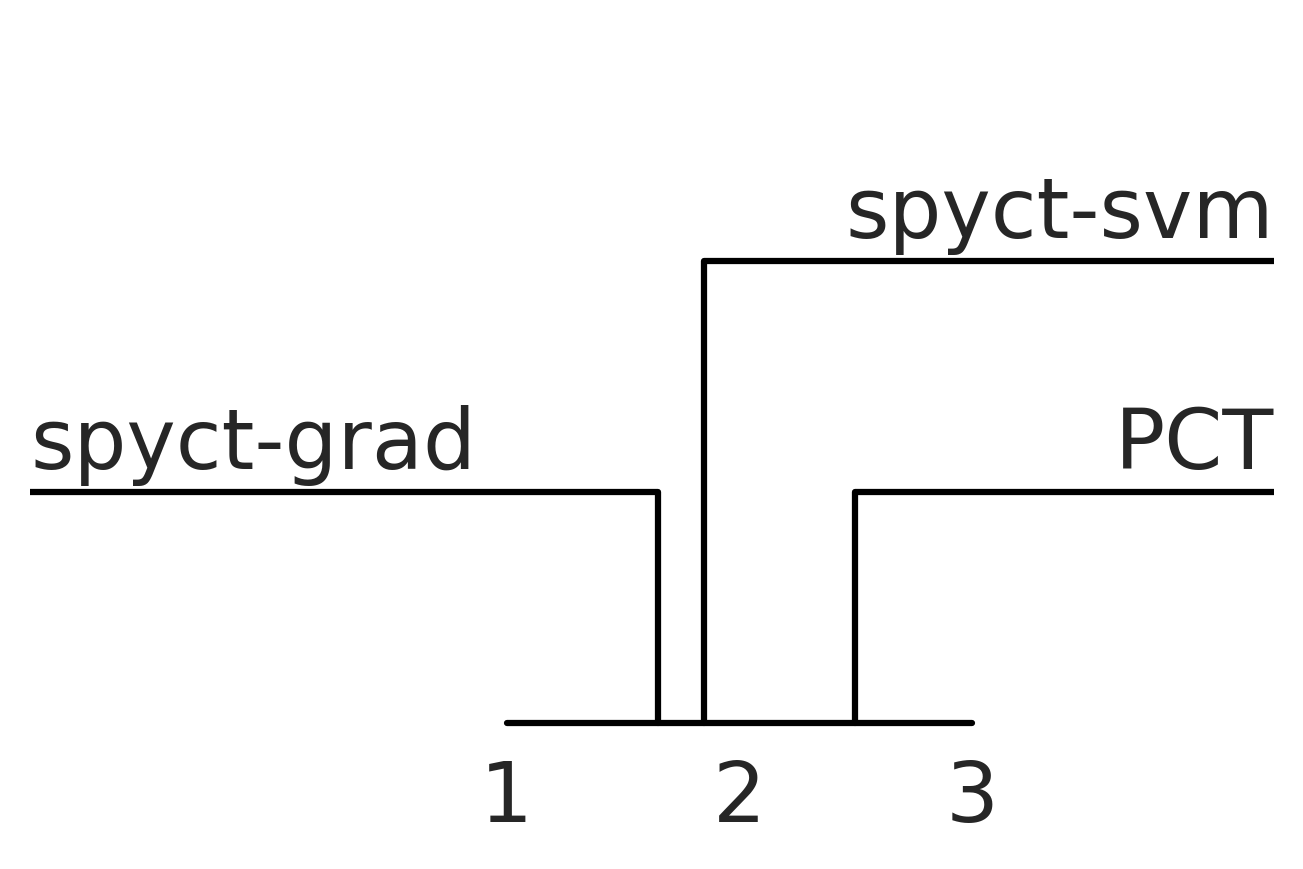} \\
    a) BIN & b) MTR 
    \end{tabular}
    \caption{Average ranks of predictive performances obtained by single-tree methods on binary classification and multi-target regression benchmark datasets.}
    \label{fig:tree_diagrams}
\end{figure}

Even though we believe the oblique PCTs are mostly suited for use in ensembles, we briefly discuss our experiments for learning single trees.
The results are illustrated in Figure~\ref{fig:tree_diagrams} by presenting the average ranks on two tasks: BIN and MTR. Note that the same behavior can be observed for all the other tasks.
As discussed in the introduction, most existing oblique trees are applicable only to binary classification problems. 
We see that our proposed methods achieve competitive performance also in the single tree setting.
The results for the \textsc{OC1} method on 3 datasets (\emph{bioresponse}, \emph{OVA\_Breast} and \emph{OVA\_Lung}) are missing, because the algorithm exited without returning a tree when it was unable to find a split of the root node. 
It was assigned the worst ranking in those cases.
We should note that even though \textsc{WODT} method performed poorly on BIN datasets, it was on par with other methods on MCC datasets.
On SOP tasks, oblique PCTs generally outperformed standard PCTs, as illustrated in Figure~\ref{fig:tree_diagrams}b.

\section{Feature importance scoring}
\label{sec:fimpresults}

\begin{figure}
    \centering
    \begin{tabular}{cc}
    \includegraphics[width=0.46\textwidth]{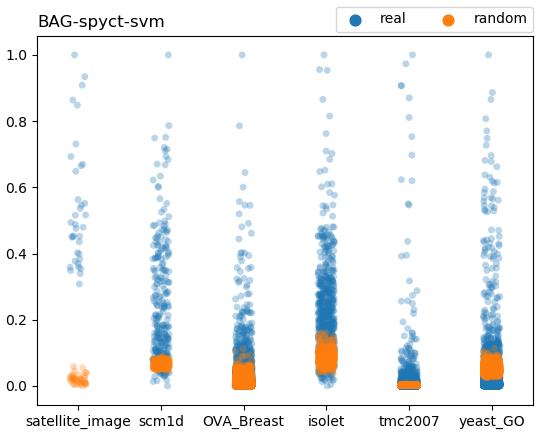} &
    \includegraphics[width=0.46\textwidth]{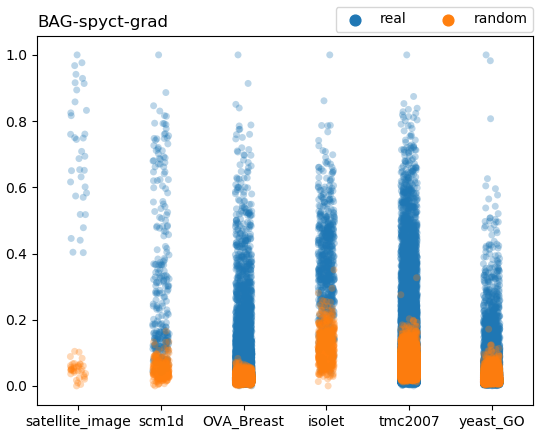} \\
    \end{tabular}
    \caption{Feature importance scores obtained on different datasets by BAG-spyct-svm (left) and BAG-spyct-grad (right) methods. Results shows that the models do not rely on irrelevant features.}
    \label{fig:fimps}
\end{figure}

We demonstrate the capability of extracting meaningful feature importance scores from the models using the approach described in Section~\ref{sec:fimp}.
To evaluate the proposed approach, we added random noise features to the datasets and then learned \textsc{BAG-spyct-svm} and \textsc{BAG-spyct-grad} models on the expanded datasets.
If the dataset originally had $d$ features, we added another $d$ features with random values, so the number of features has doubled.
We took data sparsity into account, so the ratio of non-zero values did not change (i.e., sparse datasets remained sparse).

Figure~\ref{fig:fimps} presents the results on 6 datasets -- we illustrate this on one dataset for each task.
The importances of real features have diverse values which is expected: Datasets can contain features that are informative and features that are less informative or not informative at all.
The main observation is that the added random features do not obtain high importance scores.
This indicates that obtained importance scores are meaningful, and that methods are resilient to spurious features.
This holds even on datasets with thousands of features and on datasets with sparse features.

\section{Parameter sensitivity analysis}
\label{sec:parameters}

\begin{table}
    \centering
    \begin{tabular}{l r r r r r}
    Dataset & Task & N & D & T & Sparse \\ \hline
    \emph{cholesterol} \cite{openml} & STR & 303     & 13    & 1 & \\
    \emph{pol} \cite{openml}         & STR & 15000   & 48    & 1 & \\
    \emph{qsar-30007} \cite{openml}  & STR & 534     & 1024  & 1 & \\
    
    \emph{andro} \cite{mulan}       & MTR & 49      & 30    & 6     & \\
    \emph{atp7d} \cite{mulan}       & MTR & 296     & 411   & 6     & \\
    \emph{wq} \cite{mulan}          & MTR & 1060    & 16    & 14    & \\
    
    \emph{arrhythmia} \cite{openml}       & BIN & 452     & 279   & 1   & \\
    \emph{OVA\_Endometrium} \cite{openml} & BIN & 1545    & 10935 & 1   & \\
    \emph{transfusion} \cite{openml}      & BIN & 748     & 4     & 1   & \\
    
    \emph{energy\_efficiency} \cite{openml}  & MCC & 768     & 8     & 37    & \\
    \emph{gcm} \cite{openml}                 & MCC & 190     & 160063 & 14   & \\
    \emph{gesture} \cite{openml}             & MCC & 9873    & 32    & 5     & \\
    
    \emph{birds} \cite{mulan}       & MLC & 645     & 260   & 19    & T \\
    \emph{delicious} \cite{mulan}    & MLC & 16105   & 500   & 983   & D,T \\
    \emph{mediamill} \cite{mulan}   & MLC & 43907   & 120   & 101   & T \\
    
    \emph{ara\_scop\_GO} \cite{schietgat} & HMLC & 9843 & 2003  & 572   & T \\
    \emph{imclef07a} \cite{kocev_phd}   & HMLC & 11006  & 80    & 96    & T \\
    \emph{yeast\_seq\_FUN} \cite{schietgat} & HMLC & 3932 & 478 & 594   & T \\
    \end{tabular}
    \caption{Properties of datasets for parameter sensitivity study. Columns show predictive modelling task, number of examples (N), number of features (D), number of targets (T) and whether the input or output is sparse.}
    \label{tab:param_datasets}
\end{table}

We performed parameter sensitivity analysis to investigate the influence of the different parameter values on the performance of the proposed methods.
The analysis was performed on a different selection of datasets than the benchmarking experiments, to avoid any test set leakage.
We used 3 datasets per task, yielding 18 datasets -- their properties are presented in Table~\ref{tab:param_datasets}.
For estimating the predictive performance, we used 5-fold cross validation.%, to speed up the experiments.
We experimented with different regularization strengths $C \in \{0.1, 1, 10, 100, 1000 \}$ and different numbers of ensemble members (1, 10, 25, 50, 100). 
We also tried different thresholds for impurity reduction required to accept a split: $\{ 0\%, 5\%, 10\%, 20\%\}$. 
Note that at 0\%, pre-pruning based on impurity reduction is turned off.

\begin{figure}[bt!]
    \centering
    \begin{tabular}{c}
    \includegraphics[width=0.95\textwidth]{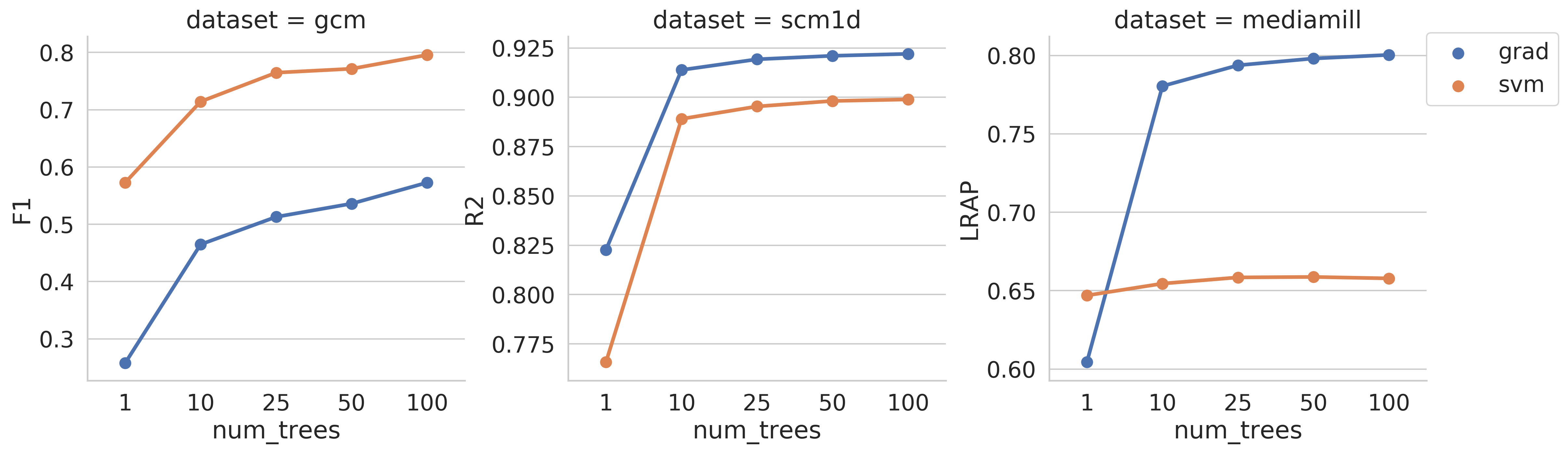} \\
    \includegraphics[width=0.95\textwidth]{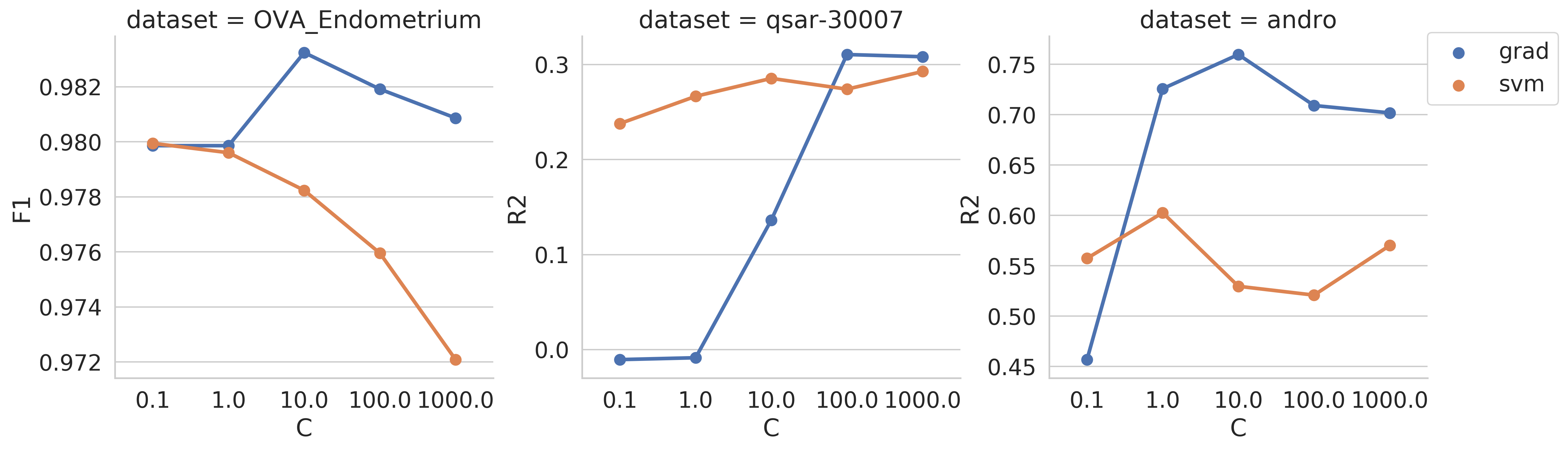} \\
    \includegraphics[width=0.95\textwidth]{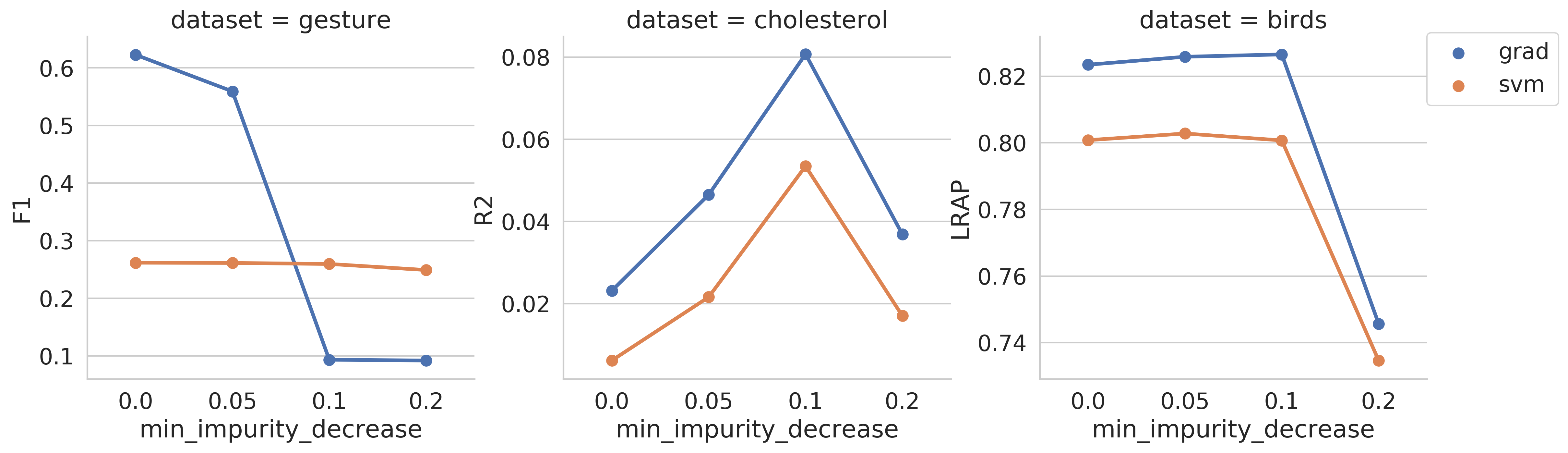} \\
    \end{tabular}
    \caption{Predictive performance obtained with different numbers of ensemble members (top), regularization strengths (middle) and impurity reduction thresholds (bottom).}
    \label{fig:params}
\end{figure}

We present the most interesting observations in Figure~\ref{fig:params}.
First, as expected, increasing the number of trees in the ensemble leads to better predictive performance.
However, on several datasets (e.g., the \emph{scm1d} dataset) the improvements become relatively small already from 10-25 trees onward.
Second, the regularization strength $C$ is difficult to get right. 
If the value is too low, the model does not fit the data enough.
If it is too high, the model can overfit and the split weights are less sparse which can lead to large model sizes (memory-wise).
Gradient descent variant appeared to be more sensitive to this parameter than the SVM variant.
The performance-wise results show that it is less damaging if $C$ is too large, compared to if it is too small.
Third, the impurity reduction threshold determines how good a split must be to be accepted (otherwise the current tree branch stops growing).
Similar to the regularization strength, if the threshold is set too high, the model does not fit the data enough.
If it is set too low, we risk overfitting the data (e.g., the \emph{cholesterol} dataset) and/or inflating the model for no benefit (increases learning time and memory requirements, e.g., the \emph{birds} dataset).
Finally, our benchmarking experiments have shown that the selected default parameter values work well overall, however, they are unlikely to be optimal for any particular dataset. The best performance on a given dataset will be obtained by tuning the parameters for that dataset.

\section{Conclusions}
\label{sec:conclusion}
In this paper, we propose two methods for learning oblique predictive clustering trees. The nodes in oblique trees contain oblique splits that have linear combinations of features in the tests, hence the splits correspond to an arbitrary hyperplane in the input space.
The first method starts by clustering the examples based on the target values and then learns a hyperplane in the input/feature space that approximates this clustering (the spyct-svm variant).
The second method uses fuzzy membership indicators and weighted variance to directly optimize the hyperplane to minimize the impurity of examples on both sides (the spyct-grad variant).
The proposed methods are designed to have drastically lower learning time on datasets with many targets compared to axis-parallel predictive clustering trees, while being capable of addressing many predictive modeling tasks, including structured output prediction.
They can also exploit the sparsity present in the input/output data to more efficiently learn the predictive models.

We evaluated the proposed methods in a single tree and ensemble settings on 60 benchmark datasets for 6 predictive modeling tasks: binary classification, multi-class classification, multi-label classification, hierarchical multi-label classification, single-target regression, and multi-target regression.
The results from the empirical study show that the predictive performance of ensembles of \textsc{spyct-svm} and \textsc{spcyt-grad} trees is on-par with state-of-the-art methods on all considered predictive modeling tasks, often exceeding it (especially \textsc{spyct-grad}).
The learning times on datasets with high dimensional input and output spaces are orders of magnitude lower than learning times of standard PCTs, especially when the data is sparse.
We also demonstrate the potential for extracting meaningful feature importance scores from the models, additionally indicating that the models are resilient to irrelevant features.

In future work, we plan to extend our work in several directions. To begin with, random forest ensembles of proposed oblique PCTs were not very successful. Hence, we plan to research the reasons behind these results and develop an improved version of the algorithm. Next, we will extend the developed methods towards the task of semi-supervised learning, where axis-parallel PCTs scale especially poorly in terms of computational cost due to a large number of clustering attributes considered there. Last but not least, we will evaluate the use of oblique trees for efficient learning of feature rankings in all of the above tasks as well as in the context of semi-supervised learning.

\section*{Acknowledgements}
This work was supported by projects from the Slovenian Research Agency through the research program grant number P2-0103 (Young Researcher grant to TS) and the project grant number J2-9230.

%\color{blue} 
%Who funded us.
%\color{black}

\bibliographystyle{elsarticle-num}
\bibliography{references}

\section*{Appendix}

\begin{figure}[bt!]
    \centering
    \includegraphics[width=0.9\textwidth]{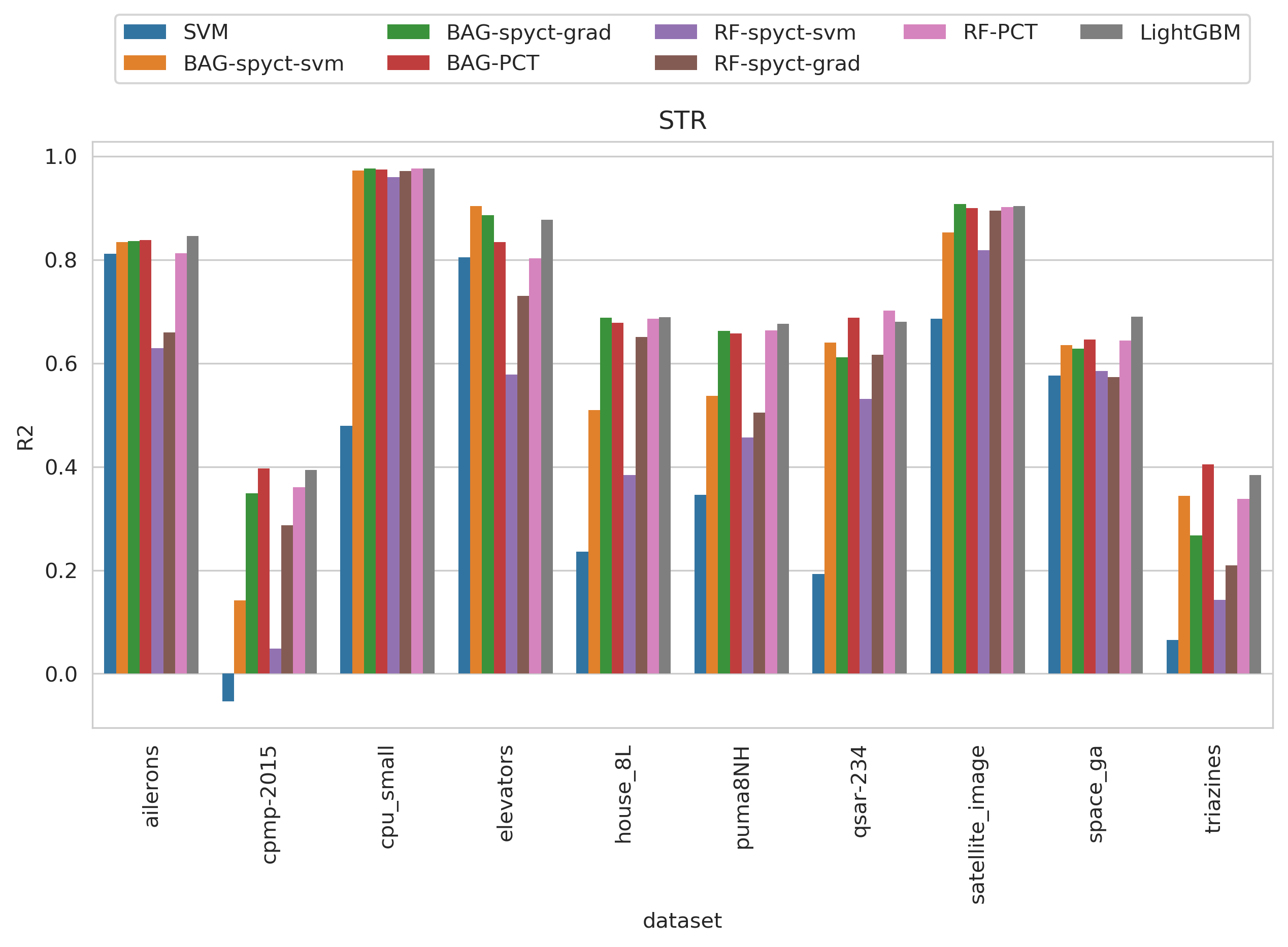}
    \caption{Predictive performance of ensemble methods and SVMs on regression datasets}
    \label{fig:perf_str_others}
\end{figure}

\begin{figure}[bt!]
    \centering
    \includegraphics[width=0.9\textwidth]{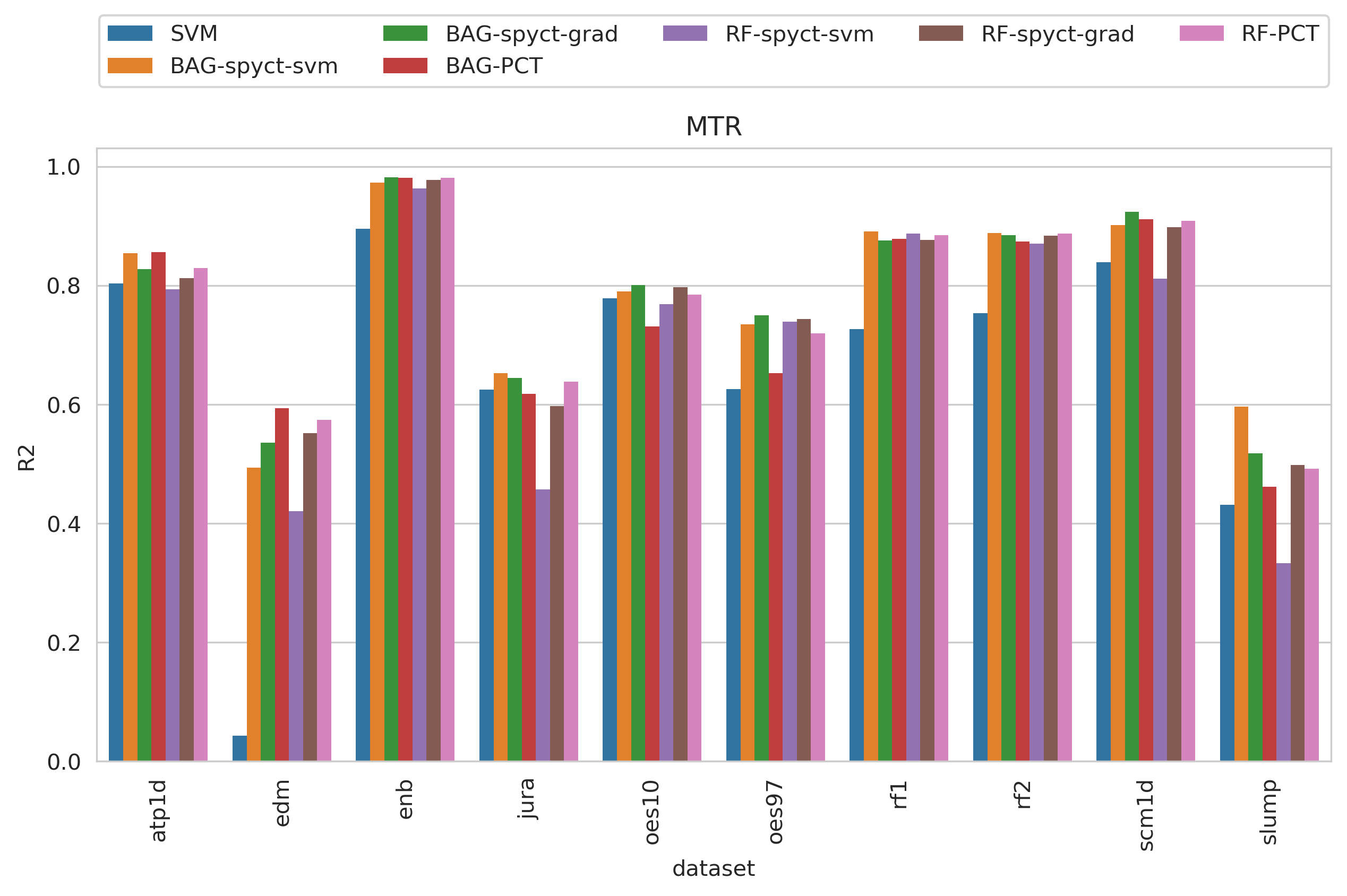}
    \caption{Predictive performance of ensemble methods and SVMs on multi-target regression datasets}
    \label{fig:perf_mtr_others}
\end{figure}

\begin{figure}[bt!]
    \centering
    \includegraphics[width=0.9\textwidth]{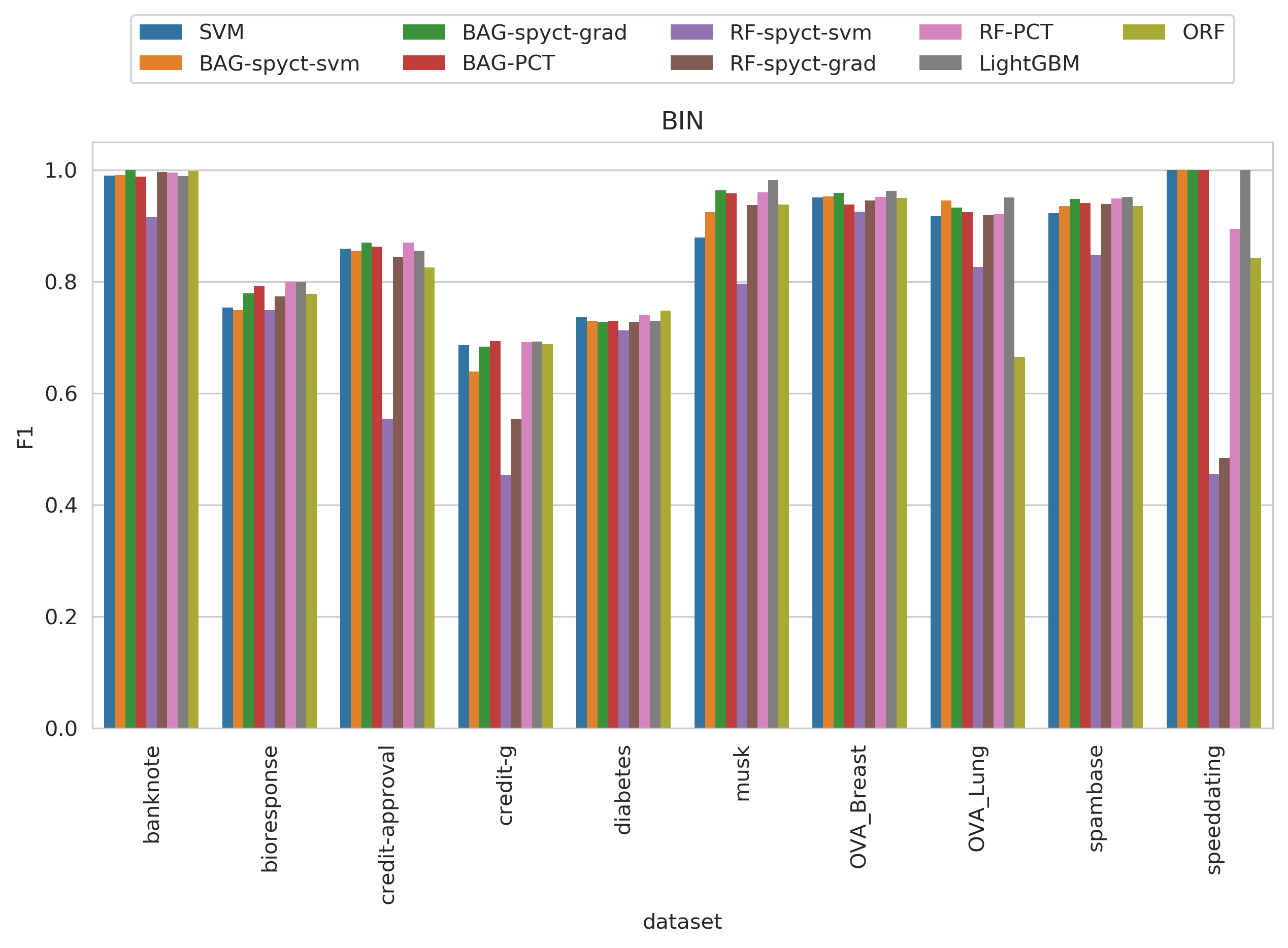}
    \caption{Predictive performance of ensemble methods and SVMs on binary classification datasets}
    \label{fig:perf_bin_others}
\end{figure}

\begin{figure}[bt!]
    \centering
    \includegraphics[width=0.9\textwidth]{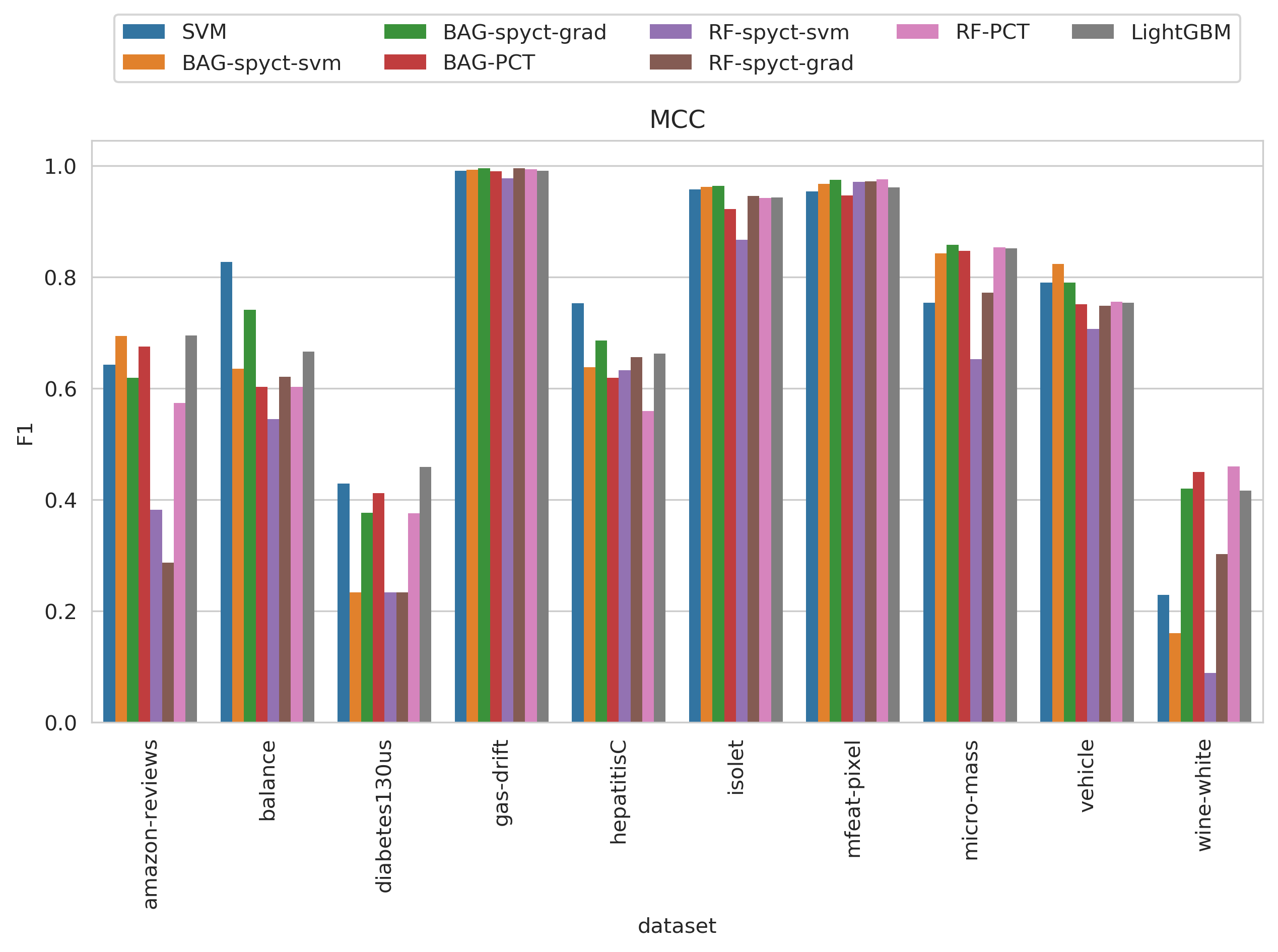}
    \caption{Predictive performance of ensemble methods and SVMs on multi-class classification datasets}
    \label{fig:perf_mcc_others}
\end{figure}

\begin{figure}[bt!]
    \centering
    \includegraphics[width=0.9\textwidth]{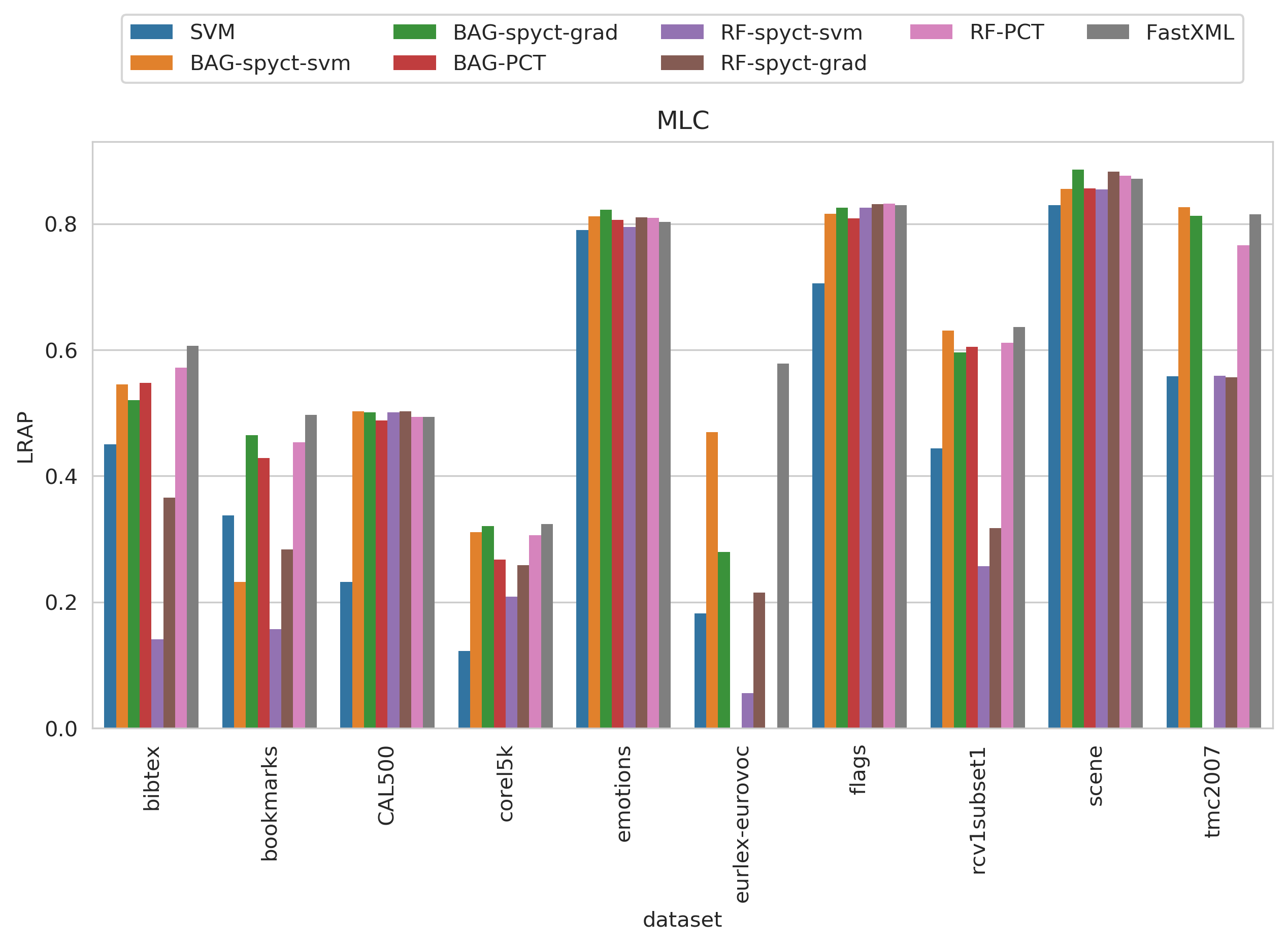}
    \caption{Predictive performance of ensemble methods and SVMs on multi-label classification datasets}
    \label{fig:perf_mlc_others}
\end{figure}

\begin{figure}[bt!]
    \centering
    \includegraphics[width=0.9\textwidth]{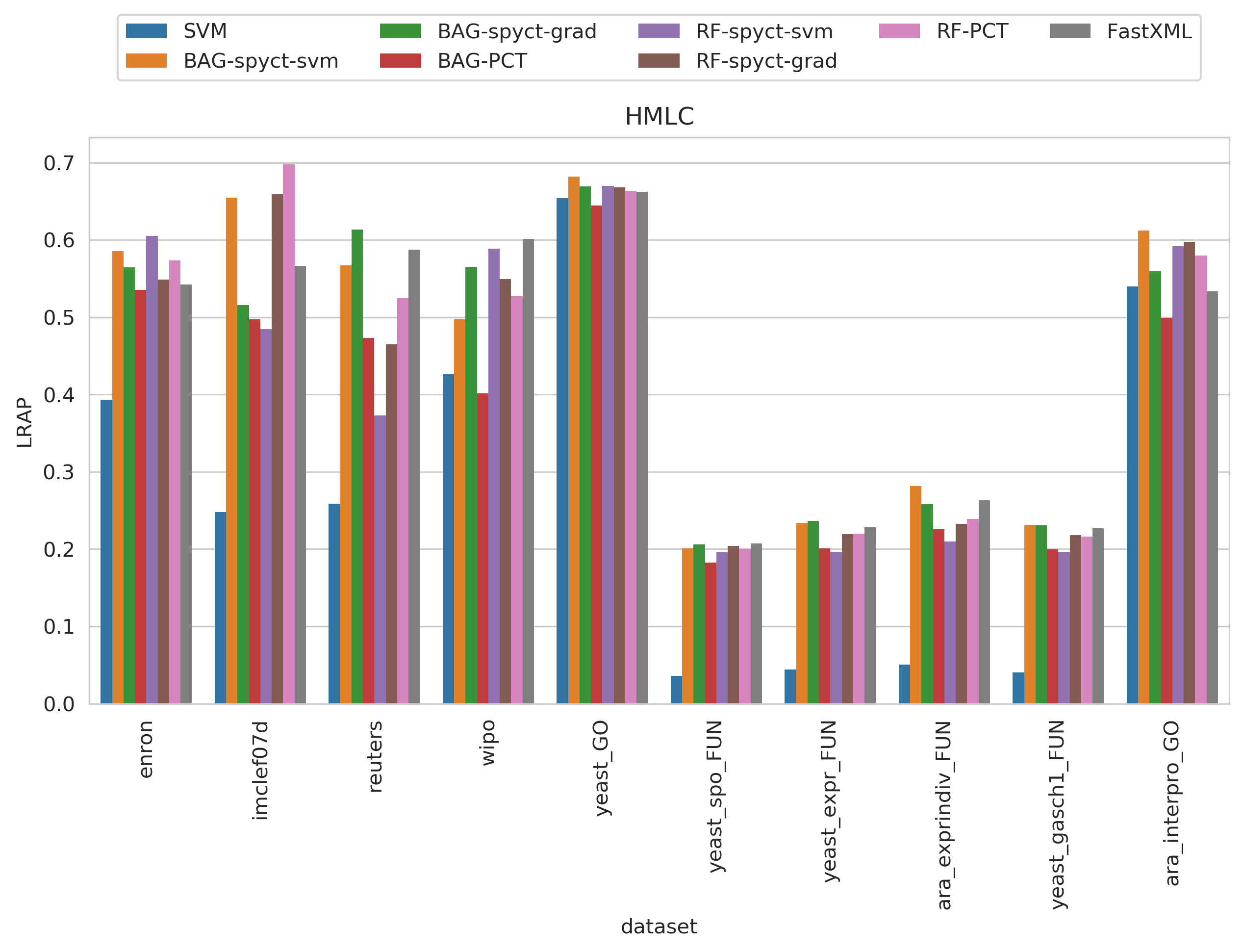}
    \caption{Predictive performance of ensemble methods and SVMs on hierarchical multi-label classification datasets}
    \label{fig:perf_hmlc_others}
\end{figure}

\end{document}